%% file: main_new.tex
\documentclass[letterpaper]{article}
\usepackage[preprint]{aaai2027}
\usepackage[hyphens]{url}
\usepackage{graphicx}
\usepackage{natbib}
\usepackage{bibunits}
\defaultbibliographystyle{aaai2027}
\usepackage{caption}
\usepackage{algorithm}
\usepackage{algorithmic}
\usepackage{booktabs}
\usepackage{amsmath}
\usepackage{placeins}
\usepackage{tikz}
\usetikzlibrary{arrows.meta,calc,fit,positioning}

\input{paper_macros}

\input{sup_setup}

\title{\methodname{}: It Looks Like CAD, but Does It Work?\\
Evaluating Parametric Design, Assembly Reasoning, and Physics Simulation}

\author{
Harmanjot Singh,
Abhra Dubey,
Jorge Alejandro Amador Herrera\corresponding
}

\affiliations{
Mohamed Bin Zayed University of Artificial Intelligence\\
Abu Dhabi, UAE\\
\{harmanjot.singh,
abhra.dubey,
jorge.herrera\}@mbzuai.ac.ae
}

\begin{document}

\maketitle

\begin{abstract}
\rev{A CAD model is not engineering-grade merely because it looks correct. It
must satisfy design requirements, respond predictably to parameter changes,
support controlled edits, match a reference structural response under a
declared analysis, and connect to other parts through valid joints. We present
\methodname{}, a two-track benchmark for these capabilities. \trackp{}
evaluates 300 parametric parts, each used for one zero-to-CAD task and one
functional-editing task (600 tasks in total), through boundary-representation
(B-Rep) validity, engineering and DFM checks, parameter-family perturbations,
functional editing, and matched linear-static FEA in CalculiX. \tracka{}
evaluates 150 body pairs through ranked joint retrieval, exact face-and-edge
grounding, joint-frame prediction, and kinematic verification. Across eight
multimodal, code-capable models, editing supplied CAD is substantially easier
than generating it, while complex edits and matched FEA remain difficult.
Assembly predictions often locate the relevant region but fail to recover the
recorded joint or mating entities. These results show that CAD evaluation must
test engineering behavior rather than appearance alone.}
\end{abstract}

\section{Introduction}

\rev{An AI-generated part can look correct even when its program fails, a
named dimension controls the wrong feature, an edit damages a mounting hole,
or its simulated response differs from the reference. An assembly prediction
can likewise identify a plausible region while selecting entities that do not
recover the recorded mating relation. Image comparison misses these failures
because CAD is both geometry and an executable engineering model.}

\rev{A parametric program maps design parameters to a B-Rep solid,
$C:\boldsymbol{\theta}\mapsto B$. Testing only
$B=C(\boldsymbol{\theta}_0)$ does not establish whether
$\boldsymbol{\theta}$ controls the intended feature. For bodies $B_A$ and
$B_B$, an assembly hypothesis $h=(j,e_A,e_B)$ must select a joint family $j$
and exact mating entities $e_A\in\mathcal E(B_A)$ and
$e_B\in\mathcal E(B_B)$. An FEA comparison must evaluate the generated and
reference responses under the same material, supports, loads, and analysis
contract. Shape similarity establishes none of these properties.}

Recent work addresses construction histories, executable programs,
editing, manufacturability, assembly, and physics-aware agents
\cite{wu2021deepcaddeepgenerativenetwork,
willis2021fusion360gallerydataset,
khan2024text2cadgeneratingsequentialcad,
xie2025texttocadquerynewparadigmcad,
dong2026musebenchmarkingmanufacturablefunctional,
son2026selfimprovingcadgenerationagents}. Existing CAD benchmarks measure
different stages with heterogeneous data and outputs. A generation score does
not reveal whether the result remains editable, \rev{recovers the recorded assembly
relation, or matches the reference structural response.}

\begin{figure*}[t]
\centering
\includegraphics[width=\textwidth]{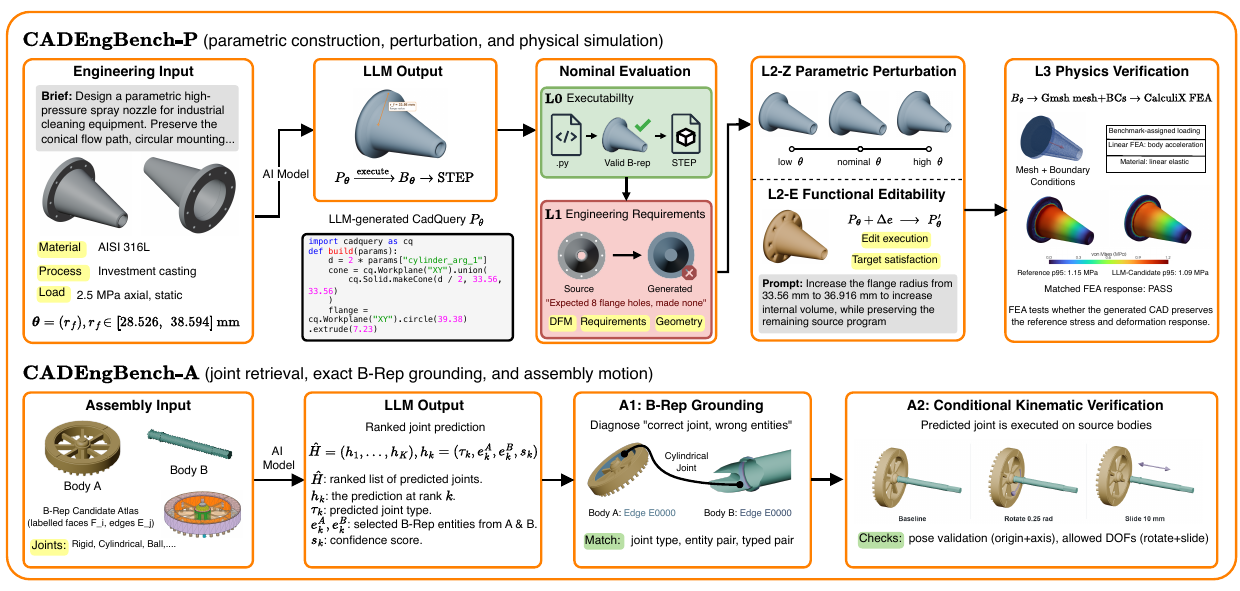}
\caption{\textbf{From generated shape to engineering checks.} \trackp{}
constructs, varies, edits, and simulates executable parts; \tracka{} grounds
relations in exact B-Rep entities and conditionally checks motion.}
\label{fig:workflow}
\end{figure*}

\rev{\methodname{} evaluates these capabilities in two separate tracks
(Figure~\ref{fig:workflow}). \trackp{} executes and measures parts, applies
controlled parameter changes and edits, and runs FEA where a reference analysis
is available. \tracka{} evaluates joint and entity retrieval, frame estimation,
and motion on supplied body pairs. Measurements, entity IDs, meshes, and solver
outputs support each verdict. The tracks remain separate: \trackp{} evaluates
single parts, while \tracka{} evaluates assembly relations and motion.}

We make four key contributions:
\begin{itemize}
    \item We develop a two-track benchmark for engineering-grade CAD:
    \trackp{} provides 600 zero-to-CAD and functional-editing tasks, while \tracka{} provides 150 assembly pairs.

    \item We develop an evaluation hierarchy that moves beyond
    nominal shape similarity to test B-Rep validity, engineering
    and DFM requirements, operational parameterization, functional edits, and
    preservation of non-target design properties.

    \item We introduce assembly evaluation in which models
    must retrieve the exact mating faces or edges, predict the joint, and demonstrate motion through kinematic simulation.

    \item We introduce FEA-based physics verification across parameter
    changes using CalculiX to compare stress, deformation,
compliance, and stress concentration across parametric design families.
\end{itemize}

\section{Related Work and Evaluation Gap}

\paragraph{CAD data and representations.}
ABC provides large-scale B-Reps, while DeepCAD and Fusion 360 Gallery made
ordered sketch--feature histories available for generation and reconstruction
\cite{koch2019abcbigcadmodel,wu2021deepcaddeepgenerativenetwork,
willis2021fusion360gallerydataset}. These works established CAD as an
executable sequence, but evaluate single parts at one final state.

\paragraph{LLM-generated and text-to-CAD.}
Text2CAD conditions construction sequences on descriptions; newer language
and vision-language systems generate executable CAD programs from text,
images, or geometric observations
\cite{khan2024text2cadgeneratingsequentialcad,
xie2025texttocadquerynewparadigmcad,
doris2025cadcoderopensourcevisionlanguagemodel,
ataei2026zerotocadagenticsynthesisinterpretable,
pyatov2026cadfsbigcadprogram,
elistratov2026cadevolvecreatingrealisticcad,
li2026highfidelitycadgenerationllmdriven}. These methods make text-to-CAD
more capable and scalable, but execution or geometric similarity does not
establish correct parameter, editing, assembly, or physical behavior.

\paragraph{CAD evaluation and editing.}
CADBench and Text2CAD-Bench broaden multimodal and text-conditioned generation;
BenchCAD introduces industrial part families and executable code editing;
CADTests checks prompt requirements through executable assertions
\cite{zhang2026benchcadcomprehensiveindustrystandardbenchmark,
doris2026cadbenchmultimodalbenchmarkaiassisted,
wang2026text2cadbenchbenchmarkllmbasedtexttoparametric,
mallis2026texttocadevaluationcadtests}. HistCAD tests whether dimensional edits
preserve sketch constraints, while MUSE emphasizes functional,
manufacturable, and assemblable specifications
\cite{dong2026histcadconstraintawareparametrichistorybased,
dong2026musebenchmarkingmanufacturablefunctional}. Interactive systems also
study tool-mediated or multimodal editing
\cite{perrett2026neuralcadeditexpertbenchmarkmultimodalinstructed,
hu2026itercaditerativemultimodalagent}. These benchmarks add engineering
evidence, but cover discrete subsets of the CAD workflow.

\paragraph{Assembly and physics.}
AutoMate and JoinABLe predict pairwise mates from B-Rep geometry; ArtiCAD,
AssemblyBench, and sequence planners evaluate articulation, assembly order, or
feasible trajectories
\cite{jones2021automatedatasetlearningapproach,
willis2022joinablelearningbottomupassembly,
shui2026articadarticulatedcadassembly,
li2026assemblybenchphysicsawareassemblycomplex,
tian2022assembleallphysicsbasedplanning,
tian2024asapautomatedsequenceplanning}.
Physics-aware CAD agents use FEA as generation or refinement feedback
\cite{ballegeer2026bendfmtaxonomysyntheticcad,
son2026selfimprovingcadgenerationagents,
berger2026physicsintheloophybridagenticarchitecture}.
These directions evaluate assembly or physics separately from executable
parametric generation.

\paragraph{Evaluation gap.}
Most CAD benchmarks stop at one boundary: one-value shape generation, local
editing, joint retrieval, or FEA feedback. They do not test whether executable
CAD survives parameter changes and edits, or whether assembly predictions
identify exact B-Rep entities and produce valid motion. As
Table~\ref{tab:prior-coverage} shows, \methodname{} connects these checks in two
complementary tracks and reports each failure separately.

\begin{table*}[t]
\centering
\scriptsize
\setlength{\tabcolsep}{3.1pt}
\renewcommand{\arraystretch}{0.92}
\begin{tabular*}{\textwidth}{@{\extracolsep{\fill}}lccccccc@{}}
\toprule
Work & Exec. & Param. & Edit & Eng./DFM &
B-Rep entities & Motion & FEA \\
\midrule
DeepCAD / Fusion 360 Gallery & E & -- & -- & -- & -- & -- & -- \\
HistCAD & E & E & E & -- & -- & -- & -- \\
BenchCAD / CADTests & E & -- & E & E & -- & -- & -- \\
CADBench / Text2CAD-Bench & E & -- & -- & -- & -- & -- & -- \\
MUSE & E & D & -- & E & D & -- & -- \\
AutoMate / JoinABLe & -- & -- & -- & -- & E & D & -- \\
ArtiCAD / AssemblyBench & E & -- & -- & -- & -- & E & -- \\
FEA-feedback CAD agents & E & -- & D & E & -- & -- & E \\
\textbf{\methodname{}} & E & E & E & E & E & E & E \\
\bottomrule
\end{tabular*}
\caption{\textbf{Evaluated capabilities in related work.} E means an evaluated
task, D means the capability is demonstrated or represented but not scored,
and -- means out of scope. Grouped rows show the union of the named works.}
\label{tab:prior-coverage}
\end{table*}

\section{Benchmark Design}

\subsection{Layered Evaluation Design}

\methodname{} follows each model output through progressively stronger
engineering checks. In \trackp{}, a generated program advances from execution
and geometry to requirements, parameter behavior, and matched FEA; functional
editing is evaluated separately on supplied CAD. In \tracka{}, ranked joint
predictions progress from exact B-Rep entity retrieval to frame and motion
verification. Layer-wise scores reveal where an apparently valid artifact
stops behaving as an engineering model.

\section{Dataset Construction}

\subsection{Compiling Source CAD into Evaluation Tasks}

\begin{figure*}[t]
\centering
\includegraphics[width=\textwidth]{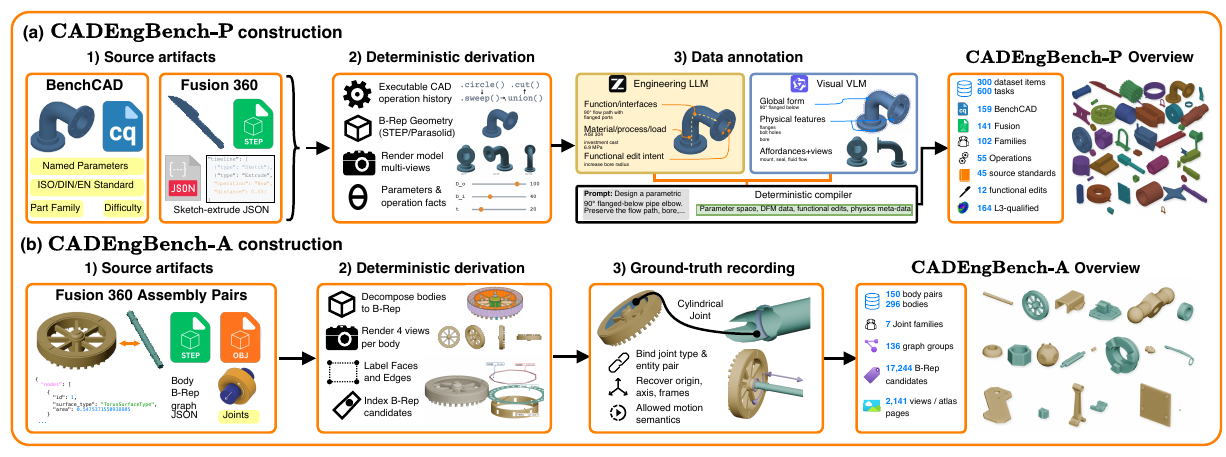}
\caption{\textbf{Dataset construction and coverage.}
\textbf{(a)} \trackp{} replays BenchCAD programs and Fusion histories, derives
B-Reps and behavioral evidence, and adds engineering/visual annotations.
\textbf{(b)} \tracka{} rebuilds Fusion body pairs, labels B-Rep candidates, and
records grounded joints, frames, and motion evidence.}
\label{fig:datasets}
\end{figure*}

\methodname{} compiles source CAD into evidence-backed tasks:
public briefs, views, and editable inputs paired with executable checks. 
CadQuery programs and Fusion histories are replayed to recover
B-Reps, parameters, operations, interfaces etc. Fusion body pairs are decomposed into B-Reps; faces and edges are indexed; and
recorded joints are bound to entities, frames, and motion
(Figure~\ref{fig:datasets}).

An engineering LLM and a VLM propose complementary functional and visual
context. A deterministic compiler retains only claims grounded in source
metadata, B-Rep measurements, or labeled benchmark scenarios, then emits
prompts and executable predicates.

\subsection{Parametric and Assembly Tracks}

\trackp{} contains 300 parts, each used for both generation and editing: 159
BenchCAD CadQuery programs
\cite{zhang2026benchcadcomprehensiveindustrystandardbenchmark} and 141 Fusion
360 Reconstruction items \cite{willis2021fusion360gallerydataset}. Their
operations span sketches and extrusions, primitives, booleans, holes etc. The set
has 100 easy, medium, and hard parts, covering 102 BenchCAD families, 55 CAD
operations, and 12 edit types. Each task includes
a brief, views, requirements, one public parameter, and hidden checks for
features that must be preserved. L0--L2 apply wherever their source checks can
be derived; 164 parts also have an L3 FEA test.

\tracka{} contains 150 body pairs from Fusion 360 assembly-joint data
\cite{jones2021automatedatasetlearningapproach} and covers
rigid, revolute, slider, cylindrical, pin-slot, planar, and ball relations.
Models receive both bodies, multiview renders, and indexed B-Rep faces and
edges. The recorded joint type, selected entities, and joint frame are hidden.
Graph, body, and source-pair separation prevents identity overlap.

\section{\trackp: Parametric Generation, Editing, and FEA}

\rev{The model converts an engineering brief, reference views, parameters,
and requirements into an executable CadQuery program. In the editing task, it
receives source CAD and a functional-edit request while specified non-target
features must remain unchanged. L0 and L1 evaluate the generated part at its
default parameters; L2-Z tests its behavior under parameter changes; L2-E tests
the functional edit and preservation of geometry; and L3 applies CalculiX FEA
to generated parts with a predefined physics test.}

\subsection{L0: Program and Solid Validity}

L0 tests whether the LLM-generated output is usable CAD. The CadQuery must execute, create at least
one valid B-Rep solid, export to STEP, and successfully re-import the exported STEP. All checks must
pass; rendering quality is not scored.

\subsection{L1: Engineering Requirements and DFM}

L1 tests whether the valid solid satisfies the engineering zero-to-CAD task. Source geometry provides
checks for dimensions, feature counts, interfaces, and item-specific
requirements. Applicable DFM checks require a 1.0\,mm minimum wall, a
2.0\,mm minimum hole diameter, and hole depth/diameter $\leq 8$. These are
general geometry screens, not manufacturing certification, and run only when
the relevant feature is present.

\subsection{L2-Z (Zero-to-CAD): Parametric Perturbations}

L2-Z tests parametric integrity: whether a declared parameter
controls the intended geometry. For item $i$, the evaluator selects one
parameter $\theta$ and a fixed set of values $\Theta_i$ containing its default,
intermediate, and valid boundary settings. It calls the same submitted
\texttt{build(params)} function at every setting.
A parameter state passes only if it produces a valid solid, satisfies its parameter-to-geometry relation, and preserves protected properties.For part $C_i(\theta)$, values
$\Theta_i$, and checks $\mathcal R_i$:
\begin{equation}
 L2Z_i=\bigwedge_{\theta\in\Theta_i}
 \left[b_i(\theta)\wedge g_i(C_i(\theta),\theta)\wedge
 \bigwedge_{r\in\mathcal R_i}r(C_i(\theta))\right],
  \label{eq:l2z}
\end{equation}
Here $b_i$ checks rebuild validity, $g_i$ checks the measured
parameter-to-geometry relation, and each $r$ checks one protected invariant.
L2-Z detects parameters that are ignored, connected
to the wrong feature, or unintentionally coupled to other dimensions.

\subsection{L2-E (Edit): Functional-Design Edits}
L2-E tests controlled modification of existing CAD rather than generation from
scratch. The LLM receives a source-native CAD artifact and one functional-edit
instruction. The edited
part $C_i^E$ must execute, achieve the requested change, and preserve every
specified non-target property:
\begin{equation}
 L2E_i=T_i(C_i^E)\wedge
 \bigwedge_{r\in\mathcal R_i^E}r(C_i^E).
 \label{eq:l2e}
\end{equation}
where $T_i$ is a hidden B-Rep measurement that verifies the requested outcome
and $\mathcal R_i^E$ contains hidden preservation checks for geometry that
should not change. Evaluation compares measured CAD properties rather than
source-code text, so a different implementation can pass if it performs the
correct edit without collateral changes.

\subsection{L3: CalculiX FEA Across Parameter Changes}

L3 compares generated and reference CAD under the same linear-static material,
supports, and loads. Its 164 parts cover linear-FEA families: axial tension (89), cantilever loading
(31), and restrained-body acceleration (44). For each matched parameter state,
the evaluator rebuilds both solids, locates the prescribed support and loading
regions, exports STEP, and creates quadratic tetrahedral meshes in Gmsh
\cite{geuzaine2009gmsh}, then solves them in CalculiX
\cite{dhondt2004finiteelement}. It rejects invalid selectors or meshes and
requires finite stress and displacement fields from the solver.

For every parameter state, L3 compares the generated and reference FEA using
95th-percentile von Mises stress, maximum displacement, and
either normalized compliance or stress concentration, depending on the load
case. For FEA quantity $q$, we measure their multiplicative difference as:
\begin{equation}
 \delta_{iq}(\theta)=
 \left|\log
 \frac{q(C_i(\theta))+\epsilon_q}
      {q(R_i(\theta))+\epsilon_q}\right|,
 \qquad
 \Delta_{iq}=\max_{\theta\in\Theta_i}\delta_{iq}(\theta),
 \label{eq:l3-response}
\end{equation}
where $C_i(\theta)$ and $R_i(\theta)$ are the generated and reference designs,
$\epsilon_q$ stabilizes values near zero, and $\Delta_{iq}$ is the worst
disagreement across the parameter family. A family passes only if every state
produces valid boundary conditions, mesh, and CalculiX solution, satisfies the
engineering limits, and remains within all FEA comparison tolerances. L3
therefore validates physically consistent CAD.

\begin{figure*}[t]
\centering
\includegraphics[width=\textwidth]{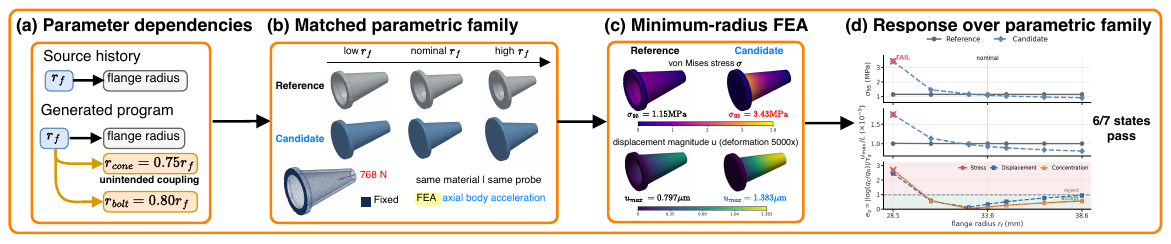}
\caption{\textbf{One parameter value can hide an incorrect dependency.}
The generated program couples cone base and bolt circle to a flange-radius
parameter (a). All states solve (b), but minimum-radius stress is $3.43$ versus
$1.15$\,MPa (c); the family trajectories expose the resulting failure (d).}
\label{fig:l3-family}
\end{figure*}

\section{\tracka: Joint Grounding and Motion}

\subsection{A0: Dataset Integrity}

A0 verifies item identity, candidate IDs, source provenance, split separation,
and that hidden joint data are absent from model inputs. It is a dataset check,
not a model score.

\subsection{A1: Joint and B-Rep Entity Retrieval}

A1 receives two bodies, multiview images, and indexed B-Rep face/edge
candidates with stable IDs. The model returns up to three hypotheses, each
containing a joint family, one entity from Body A, one from Body B, and a
confidence. A strict JSON parser rejects unknown or wrong-body IDs; exact
role-aware matching compares each hypothesis with accepted source entity pairs.
Typed match requires the joint family and entity pair in the same hypothesis.
Equivalent source entities count as correct; a miss means the recorded relation
was not retrieved.

\subsection{A2: Joint Frame and Motion}

A2 conditions on the rank-1 A1 hypothesis and predicts its joint origin and
family-specific axes or directions in Body A's coordinate frame. The origin
must match the source within 1\% of body size and each direction within
$5^\circ$. The frame becomes a URDF joint executed in PyBullet; compound
revolute/prismatic constraints represent cylindrical, planar, and pin-slot
motion. Scripted tests require allowed motion without forbidden translation or
rotation. Hidden annotations never repair an incorrect A1 prediction.

\section{Evaluation Metrics}

\paragraph{\trackp{}.}
\textbf{L0} measures executable, valid CAD.
\textbf{L1} additionally requires all applicable engineering and DFM checks.
\textbf{L2-Z} tests whether parameter changes rebuild correctly, modify the
intended feature, and preserve protected geometry.
\textbf{L2-E} requires a requested functional edit without unintended changes.
\textbf{L3 reach} records whether matched generated/reference FEA can be run,
while \textbf{L3 pair pass} additionally requires valid simulation and
agreement in structural behavior.

\paragraph{\tracka{}.}
\textbf{Valid} checks the output schema and entity identifiers.
\textbf{Entity@$k$} requires the correct B-Rep entity pair within the top $k$
predictions; \textbf{Typed@$k$} also requires the correct joint family and body
ordering. \textbf{MRR} (Mean Reciprocal Rank) rewards earlier fully correct
predictions. \textbf{A2 E2E} (End-to-End) requires a correct rank-1 joint, joint frame, and
allowed and blocked motion. The two tracks are reported separately.

\section{Experimental Protocol}

We evaluate eight multimodal, code-capable models: GPT-5.2
\cite{openai2025gpt52}, Claude Sonnet 4.5 \cite{anthropic2025sonnet45},
Gemini 3 Flash \cite{google2025gemini3flash}, GLM-4.6V
\cite{zai2025glm46v}, Kimi K2.5 \cite{moonshot2026kimik25},
Mistral Medium 3.5 \cite{mistral2026medium35}, Llama 4 Maverick
\cite{meta2025llama4}, and Qwen3.5-35B-A3B \cite{qwen2026qwen35}.
Each model receives identical track-specific inputs and produces one response
per task without retries.

In \trackp{}, L0, L1, and L2-E evaluate all 300 items; L2-Z reports the
evaluable denominator for each model (287--300). L3 evaluates 164 eligible
parts across available parameter states, including complete seven-state
families for 18 parts. In \tracka{}, 30 items were used for evaluator
development and excluded from comparison; all reported results use the
remaining 120 items.

Confidence intervals use 10,000 item-clustered bootstrap samples. Pairwise
L0--L2 comparisons use exact McNemar tests with Holm correction. Model
identifiers, inference settings, confidence intervals, and detailed result
slices are provided in the supplement.

\begin{table*}[t]
\centering
\footnotesize
\setlength{\tabcolsep}{2.6pt}
\renewcommand{\arraystretch}{0.96}
\textbf{(a) Parametric construction, editing, and physics (\trackp)}\\[2pt]
\begin{tabular*}{\textwidth}{@{\extracolsep{\fill}}lcccccc@{}}
\toprule
& \multicolumn{3}{c}{Zero-to-CAD} & \multicolumn{1}{c}{Edit} &
\multicolumn{2}{c}{L3 physics} \\
\cmidrule(lr){2-4}\cmidrule(lr){5-5}\cmidrule(lr){6-7}
System & L0 (\%) & L1 (\%) & L2-Z (\%; $N_s$) & L2-E (\%) &
Reach (\%; parts/164) & Pair pass (\%; pass/$N_s$) \\
\midrule
GPT-5.2          & 41.3 & 22.0 & 31.6 (291) & 70.3 & 37.8 (62/164) & 32.8 (65/198) \\
Claude 4.5       & \textbf{58.0} & 28.3 & \textbf{41.4} (290) & 66.7 & \textbf{54.3} (89/164) & 35.8 (101/282) \\
Gemini 3 Flash   & 52.7 & \textbf{30.0} & 39.8 (289) & \textbf{72.3} & 46.3 (76/164) & \textbf{46.3} (119/257) \\
GLM-4.6V         & 47.3 & 13.3 & 30.7 (287) & 59.3 & 44.5 (73/164) & 27.0 (64/237) \\
Kimi K2.5        & 53.3 & 24.3 & 41.0 (288) & 67.0 & 51.8 (85/164) & 34.8 (94/270) \\
Mistral 3.5      & 29.3 &  8.3 & 16.5 (297) & 68.3 & 26.8 (44/164) & 34.5 (49/142) \\
Llama 4 Maverick & 48.0 & 12.7 & 34.0 (288) & 60.7 & 47.6 (78/164) & 27.5 (71/258) \\
Qwen3.5-35B      & 13.3 &  5.0 &  6.3 (300) & 61.7 & 15.9 (26/164) & 36.6 (30/82) \\
\bottomrule
\end{tabular*}

\vspace{5pt}
\textbf{(b) Assembly relation and geometric grounding (\tracka)}\\[2pt]
\begin{tabular*}{\textwidth}{@{\extracolsep{\fill}}lcccccc@{}}
\toprule
& \multicolumn{1}{c}{Output} & \multicolumn{4}{c}{Grounded retrieval} &
\multicolumn{1}{c}{Kinematics} \\
\cmidrule(lr){2-2}\cmidrule(lr){3-6}\cmidrule(lr){7-7}
System & Valid (\%) & Entity@1 (\%) & Typed@1 (\%) & Typed@3 (\%) &
MRR & A2 E2E (\%; pass/120) \\
\midrule
GPT-5.2          & 100.0 & 61.7 & 12.5 & 30.0 & .197 & 11.7 (14) \\
Claude 4.5       & 100.0 & 57.5 & 14.2 & 32.5 & .221 & 12.5 (15) \\
Gemini 3 Flash   & 100.0 & \textbf{62.5} & \textbf{23.3} & \textbf{41.7} & \textbf{.310} & \textbf{15.8} (19) \\
GLM-4.6V         &  99.2 & 50.0 & 10.0 & 27.5 & .174 &  8.3 (10) \\
Kimi K2.5        &  78.3 & 56.7 & 15.0 & 32.5 & .228 &  7.5 (9) \\
Mistral 3.5      & 100.0 & 52.5 & 15.8 & 25.8 & .200 & 14.2 (17) \\
Llama 4 Maverick & 100.0 & 55.8 & 14.2 & 26.7 & .192 & 10.8 (13) \\
Qwen3.5-35B      & 100.0 & 55.0 & 14.2 & 24.2 & .190 & 10.8 (13) \\
\bottomrule
\end{tabular*}

\caption{\textbf{Primary model results.} P uses 300 tasks except L2-Z, where
$N_s$ is the number of evaluable items. L3 reach uses 164 parts; pair pass uses
only comparable generated/reference state pairs. A uses 120
tasks per system; A2 E2E includes all stages. Entity ignores joint type, while
Typed requires the recorded joint family and body roles}
\label{tab:results}
\end{table*}

\section{Results and Analysis}

\textbf{Capability profile.}
We benchmark eight leading multimodal models across parametric CAD generation,
functional editing, FEA-based physics verification, and assembly reasoning.
Table~\ref{tab:results} reveals substantial variation across both models and
evaluation stages, with no model performing best throughout the benchmark.
Gemini leads engineering-requirement satisfaction, functional editing, matched
FEA, and assembly retrieval, whereas Claude leads executability, parametric
behavior, and L3 simulation reach.

The contrasts are sharper beyond the leading systems. Qwen reaches the fewest
L3 cases but has the second-highest FEA pass rate once a comparable simulation
exists. Mistral ranks seventh on L2-Z but second on A2, while Kimi ranks second
and last, respectively. Consequently, generation and editing ranks are only
weakly associated ($\rho_s=0.214$), and parametric construction and assembly
motion ranks are nearly unrelated ($\rho_s=0.048$).

We therefore analyze where performance separates: first \trackp{}
construction and editing, then physics verification, and finally \tracka{}
grounding and motion.

\begin{center}
\centering
\includegraphics[width=0.80\columnwidth]{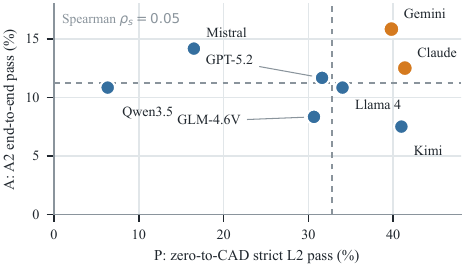}
\captionof{figure}{\textbf{Cross-track model performance.}
A near-zero rank association shows that parametric construction and assembly motion
favor different models.}
\label{fig:cross-track-map}
\end{center}

\subsection{Generation and Editing}

\textbf{Execution as an entry condition.} Across all eight systems,
1,030/2,400 generated programs pass L0, but only 432 of those also pass L1.
Thus, 58.1\% of executable outputs violate at least one stated engineering or
DFM requirement. Claude leads L0 and L2-Z, while Gemini leads L1, but
Holm-corrected McNemar tests do not separate either point estimate from the
next-best system. Executability therefore substantially overstates how often
generated CAD satisfies its engineering brief.

\textbf{CAD Editing.} Among 2,330 item--model pairs scoreable under both protocols, 1,071
pass only editing, compared with 231 that pass only generation; 468 pass both
and 560 pass neither. The 4.64$\times$ edit-only asymmetry explains the
25.5--55.4 point gaps in Figure~\ref{fig:protocol-gap}, but it is not uniform
over edit structure.

\begin{center}
\centering
\includegraphics[width=0.92\columnwidth]{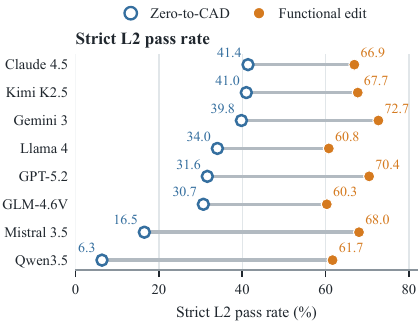}
\captionof{figure}{\textbf{Editing and generation comparison.}
Rates use only items that can be scored for both tasks.}
\label{fig:protocol-gap}
\end{center}

Fusion 360 items are grouped by their
\emph{construction-history structure} (Figure~\ref{fig:fusion-edit-structure}), not by the primitive being edited.
The five groups record whether the source creates one or multiple
bodies and whether subsequent features join or cut existing solids. The actual
edited quantity remains either an extrusion distance or a sketch-profile
radius.

A single independent-body addition passes in 247/248 cases, whereas histories
involving joins, cuts, or multiple bodies pass in only 40.2--45.8\% of cases (Figure~\ref{fig:fusion-edit-structure}). Editability, therefore, deteriorates strongly when an edit becomes structurally coupled to the existing
CAD model.. Functional-edit L2 also falls from
these slices show that the remaining difficulty is not merely parsing an edit
request; it is changing operation history while preserving dependent geometry.
Aggregate edit scores can consequently hide a large gap between isolated
feature insertion and edits that interact with existing construction history.

\begin{center}
\centering
\includegraphics[width=0.85\columnwidth]{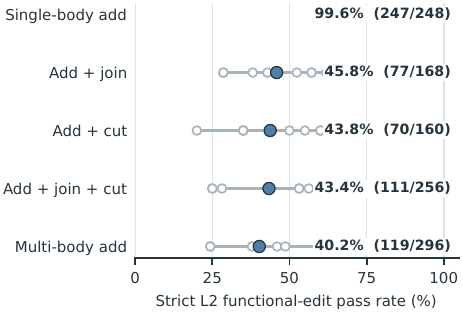}
\captionof{figure}{\textbf{Functional editing breaking.} Filled dots pool item--model outcomes; open dots show individual
systems.}
\label{fig:fusion-edit-structure}
\end{center}

\subsection{Physics Verification}

\begin{figure}[t]
\centering
\includegraphics[width=0.80\columnwidth]{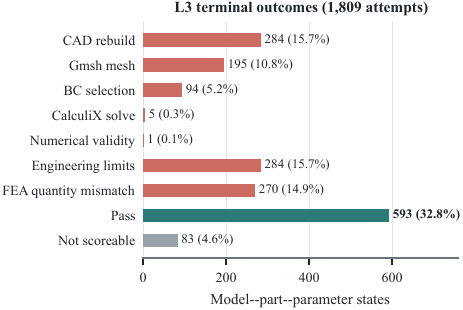}
\caption{\textbf{L3 failures at every FEA stage.}
Each of 1,809 model--part--parameter-state attempts is assigned its first
terminal outcome; 83 gray cases lack a comparable generated/reference pair.}
\label{fig:l3-terminal-outcomes}
\end{figure}

Critically, a successful solve does not establish correct physics. Models that
produce more valid CAD by passing all previous evaluation layers reach more L3 cases ($\rho_s=0.976$), but reach is
unrelated to agreement with the reference FEA ($\rho_s=0.024$). CalculiX
itself fails in only 5 of 1,809 attempts. In contrast, 573 attempts fail before
the solve because meshing or boundary-condition assignment fails, and 554
solve but violate engineering limits or disagree with the reference stress or
deformation. L3 therefore tests the complete simulation and its structural
result, not solver completion alone. Moreover, 22/435 matched cases change
verdict away from the default parameter value, so testing only one design
state can hide physics failures.

\subsection{Complex Assembly Behaviour}

\begin{center}
\centering
\includegraphics[width=0.80\columnwidth]{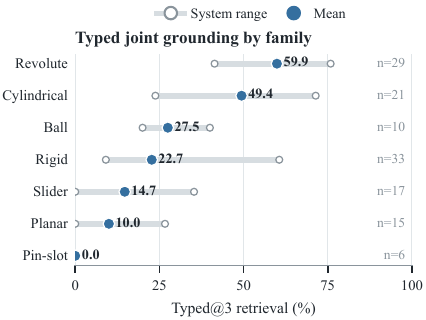}
\captionof{figure}{\textbf{Joint grounding varies by joint family.} Dots show
mean Typed@3; lines show the range across systems. Family counts $n$ may overlap because a pair may contain multiple recorded joint families.}
\label{fig:a-joint-families}
\end{center}

We also found that ill-formed predictions often describe the wrong connection. A
correct B-Rep entity pair is ranked first in 542/960 requests, but only 143
also identify the joint family and body ordering. Thus, just 26.4\% of correct
rank-1 entity localizations specify the full joint. Revolute and cylindrical
joints are easier than slider, planar, and pin-slot joints
(Figure~\ref{fig:a-joint-families}).

The joint frame is a further bottleneck: 110 of 142 correct rank-1 joints place
it within tolerance, and only 11.5\% of all requests pass end to end. Every
correctly specified ideal joint completes its requested motion, indicating
that errors arise mainly in entity selection, joint classification, and frame
placement rather than motion execution.

Revolute, cylindrical, and ball joints score highest because each resolves to one geometrically distinctive mating surface (a shaft in a bore, a spherical contact), making both localization and typing easy. Rigid joints score lower despite having zero degrees of freedom: two rigidly connected bodies often admit several plausible flat mating faces, so a model can correctly infer no relative motion, yet still select a different face pair than ground truth, failing the stricter Typed metric. Pin-slot falls to 0\% for a different reason: it is a compound constraint coupling rotation with translation along a slot axis, so a model attending to only one half of the motion misclassifies it as revolute or slider, a ``correct entity, wrong type'' error consistent with the Entity-to-Typed collapse reported above. With the smallest sample of any family ($n = 6$), a handful of such errors is enough to zero out the score.

\section{Limitations and Future Work}

Our benchmarking framework is limited by the information available in its source CAD.
Materials, tolerances, operating conditions, and design intent are sometimes
incomplete; LLM annotations cannot make them authoritative.
Consequently, L1 evaluates measurable engineering and DFM requirements rather
than certifying manufacturability. \trackp{} also tests one parameter and one
bounded edit per item. Future versions should incorporate expert-annotated
requirements, coupled parameters, and sequential edits.

\tracka{} evaluates recorded joints between body pairs rather than complete
assemblies. It therefore does not measure whether a model can recover an entire
assembly structure. A natural extension is to organize verified pairwise
relations into multi-body assembly graphs and evaluate graph-level joint
consistency and connectivity.

L3 covers standardized linear-FEA of isolated parts.
Future work should expand its physics coverage to buckling, thermal and
nonlinear analysis, contact and assembly-level mechanics, and fluid or coupled
multiphysics problems.

\section{Conclusion}

\rev{\methodname{} evaluates CAD beyond appearance and executability through
engineering requirements, parameter behavior, controlled edits, matched FEA,
and assembly grounding. Across the eight evaluated systems, performance is only
weakly aligned across these capabilities; no single score describes the full
profile.}

\rev{The benchmark exposes failures that surface checks miss: executable code
can violate design intent, generated parameter families can diverge from the
reference structural response, and plausible assembly predictions can fail to
recover the recorded mating relation. \methodname{} therefore treats CAD as an
engineering artifact rather than a plausible shape.}

\bibliography{references}

\clearpage
\onecolumn
\appendix
\setcounter{secnumdepth}{2}
\setcounter{tocdepth}{1}
\setcounter{figure}{0}
\setcounter{table}{0}
\setcounter{equation}{0}
\setcounter{algorithm}{0}
\renewcommand{\thesection}{\Alph{section}}
\providecommand{\theHfigure}{}
\providecommand{\theHtable}{}
\providecommand{\theHequation}{}
\providecommand{\theHalgorithm}{}
\renewcommand{\theHfigure}{supp.figure.\arabic{figure}}
\renewcommand{\theHtable}{supp.table.\arabic{table}}
\renewcommand{\theHequation}{supp.equation.\arabic{equation}}
\renewcommand{\theHalgorithm}{supp.algorithm.\arabic{algorithm}}
\raggedbottom
\setlength{\emergencystretch}{1.5em}
\setlength{\textfloatsep}{9pt plus 2pt minus 2pt}
\setlength{\floatsep}{8pt plus 2pt minus 2pt}
\setlength{\intextsep}{8pt plus 2pt minus 2pt}

\begin{center}
{\Large\bfseries Technical Supplement}
\end{center}
\vspace{0.5em}

\begin{bibunit}[aaai2027]
\renewcommand{\bibliography}[1]{\putbib[#1]}
\input{sup2}
\end{bibunit}

\end{document}

%% file: paper_macros.tex
\newcommand{\methodname}{\textsc{CADEngBench}}
\newcommand{\trackp}{\textsc{CADEngBench-P}}
\newcommand{\tracka}{\textsc{CADEngBench-A}}

\newif\ifshowrev
\showrevfalse
\DeclareRobustCommand{\rev}[1]{\ifshowrev{\color{blue}#1}\else{#1}\fi}

%% file: sup_setup.tex
\usepackage{xcolor}
\usepackage{listings}
\usepackage{array}
\usepackage{multirow}
\usepackage{amssymb}
\usepackage{enumitem}
\usepackage{multicol}
\usepackage{tcolorbox}
\tcbuselibrary{breakable,skins}

\definecolor{cebblue}{HTML}{356FA3}
\definecolor{cebteal}{HTML}{2A8C82}
\definecolor{ceborange}{HTML}{D97714}
\definecolor{cebcoral}{HTML}{C84D5A}
\definecolor{cebgray}{HTML}{5F6973}
\definecolor{cebcarddark}{HTML}{5A5D60}
\definecolor{cebcardline}{HTML}{AEB4B9}
\definecolor{ceblight}{HTML}{F3F5F7}
\definecolor{codegreen}{HTML}{287A50}

\lstdefinestyle{cebcode}{
  basicstyle=\ttfamily\scriptsize,
  keywordstyle=\color{cebblue}\bfseries,
  stringstyle=\color{codegreen},
  commentstyle=\color{cebgray},
  showstringspaces=false,
  breaklines=true,
  breakatwhitespace=false,
  frame=single,
  rulecolor=\color{black!20},
  backgroundcolor=\color{ceblight},
  columns=fullflexible,
  keepspaces=true,
  aboveskip=4pt,
  belowskip=4pt,
  tabsize=2
}
\lstdefinestyle{cebprompt}{
  basicstyle=\ttfamily\fontsize{6.15}{7.15}\selectfont,
  keywordstyle=\color{black},
  stringstyle=\color{black},
  commentstyle=\color{black},
  showstringspaces=false,
  breaklines=true,
  breakatwhitespace=false,
  breakindent=0pt,
  breakautoindent=false,
  frame=none,
  columns=fullflexible,
  keepspaces=true,
  xleftmargin=0pt,
  xrightmargin=0pt,
  aboveskip=1pt,
  belowskip=1pt,
  tabsize=2
}
\lstdefinestyle{cebpromptcompact}{
  style=cebprompt,
  basicstyle=\ttfamily\fontsize{5.75}{6.65}\selectfont
}
\lstdefinestyle{cebpromptmicro}{
  style=cebprompt,
  basicstyle=\ttfamily\fontsize{4.75}{5.45}\selectfont
}
\lstdefinestyle{cebassemblyprompt}{
  style=cebprompt,
  basicstyle=\ttfamily\fontsize{5.05}{5.75}\selectfont
}
\lstdefinestyle{cebcodecard}{
  basicstyle=\ttfamily\scriptsize,
  keywordstyle=\color{cebblue}\bfseries,
  stringstyle=\color{codegreen},
  commentstyle=\color{cebgray},
  showstringspaces=false,
  breaklines=true,
  breakatwhitespace=false,
  frame=none,
  columns=fullflexible,
  keepspaces=true,
  xleftmargin=0pt,
  xrightmargin=0pt,
  aboveskip=1pt,
  belowskip=1pt,
  tabsize=2
}
\lstdefinestyle{cebinterfacecode}{
  style=cebcodecard,
  basicstyle=\ttfamily\fontsize{6.15}{7.15}\selectfont,
  breaklines=false,
  columns=fullflexible,
  keepspaces=true,
  tabsize=4
}
\newcommand{\passmark}{\textcolor{codegreen}{\textbf{PASS}}}
\newcommand{\failmark}{\textcolor{cebcoral}{\textbf{FAIL}}}
\newcommand{\nrmark}{\textcolor{cebgray}{\textbf{N/R}}}

\newenvironment{cebcard}[2][cebcarddark]{%
  \begin{tcolorbox}[
    enhanced,
    title={#2},
    title after break={#2 (continued)},
    colback=#1!7!white,
    colframe=#1!70!black,
    colbacktitle=#1!92!black,
    coltitle=white,
    coltext=black,
    fonttitle=\bfseries,
    boxrule=0.55pt,
    arc=5pt,
    outer arc=5pt,
    left=7pt,
    right=7pt,
    top=8pt,
    bottom=6pt,
    toptitle=3pt,
    bottomtitle=3pt,
    before skip=6pt,
    after skip=6pt,
    before upper={\lstset{backgroundcolor=\color{#1!7!white}}},
    breakable
  ]
}{%
  \end{tcolorbox}
}

%% file: sup2.tex

\begin{abstract}
This supplement provides the full methodological and experimental details for
\methodname{}. It describes dataset selection and annotation, model inputs and
outputs, layer-specific scoring rules, evaluation methodologies,
model configurations, uncertainty estimates, and worked examples.
\end{abstract}

\FloatBarrier
\section{Dataset Construction}
\label{sec:data}

\subsection{Source Datasets and Task Units}

\trackp{} contains two source-derived subsets totaling 300 single parts. Each part defines one zero-to-CAD request and one
separate functional-edit request, yielding 600 tasks. \trackp{}-BenchCAD is a 159-part selection from BenchCAD that
provides executable CadQuery programs \citep{zhang2026benchcadcomprehensiveindustrystandardbenchmark}, whereas the \trackp{}-Fusion360 is a 141-part selection from the Reconstruction subset of the Autodesk Fusion 360 Gallery Dataset that provides replayable construction histories and B-Rep geometry \citep{willis2021fusion360gallerydataset}. The two sources therefore, share an
executable geometric evaluation but retain source-native editing representations.

\tracka{} is derived from the Autodesk Fusion 360 Gallery Assembly-Joint dataset, release \texttt{j1.0.0}, introduced with AutoMate \citep{jones2021automatedatasetlearningapproach}. We select 150 assembly body pairs with source-recorded joint
relations, B-Rep graphs, candidate face and edge labels, joint frames, and kinematic motion semantics. Thirty A pairs were used
to test prompts, parsers, and evaluator execution. They were excluded from
model comparison; all reported model comparisons use the remaining 120 pairs.

\begin{figure}[t]
\centering
\includegraphics[width=\textwidth]{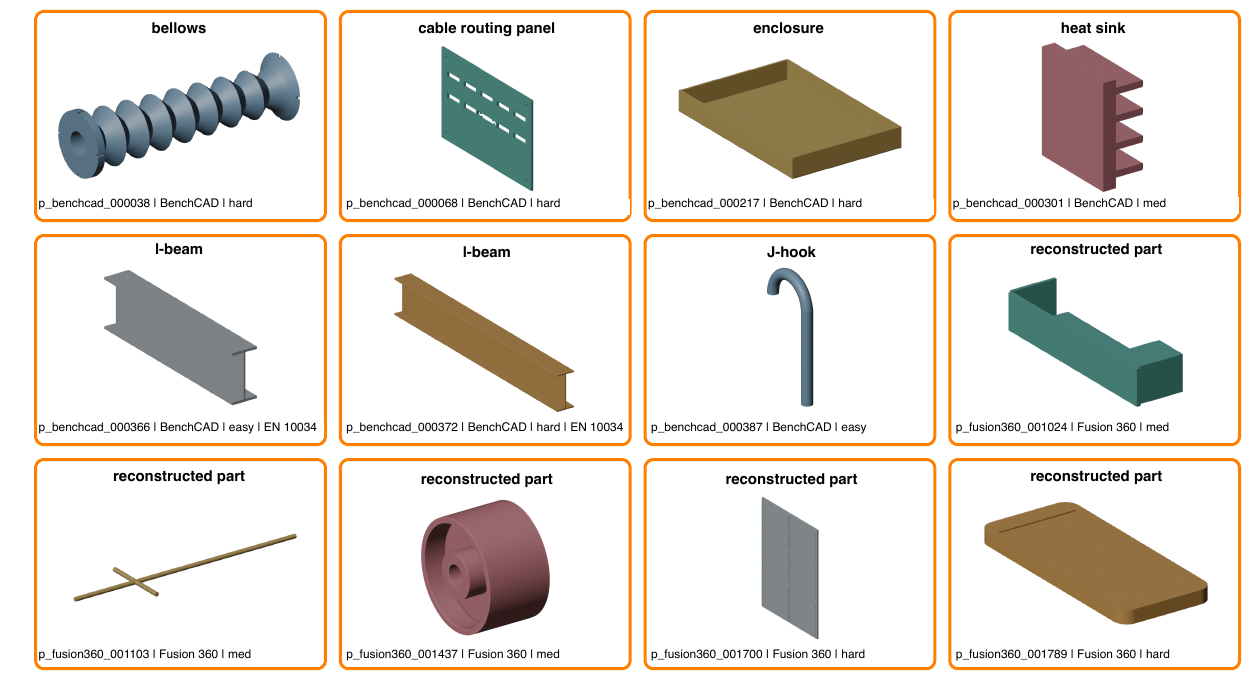}
\caption{\textbf{Additional \trackp{} examples.} The gallery contains
operation-rich BenchCAD parts and parts reconstructed from Fusion 360 histories.
All renders use source B-Reps}
\label{fig:p-gallery}
\end{figure}

\begin{figure}[t]
\centering
\includegraphics[width=\textwidth]{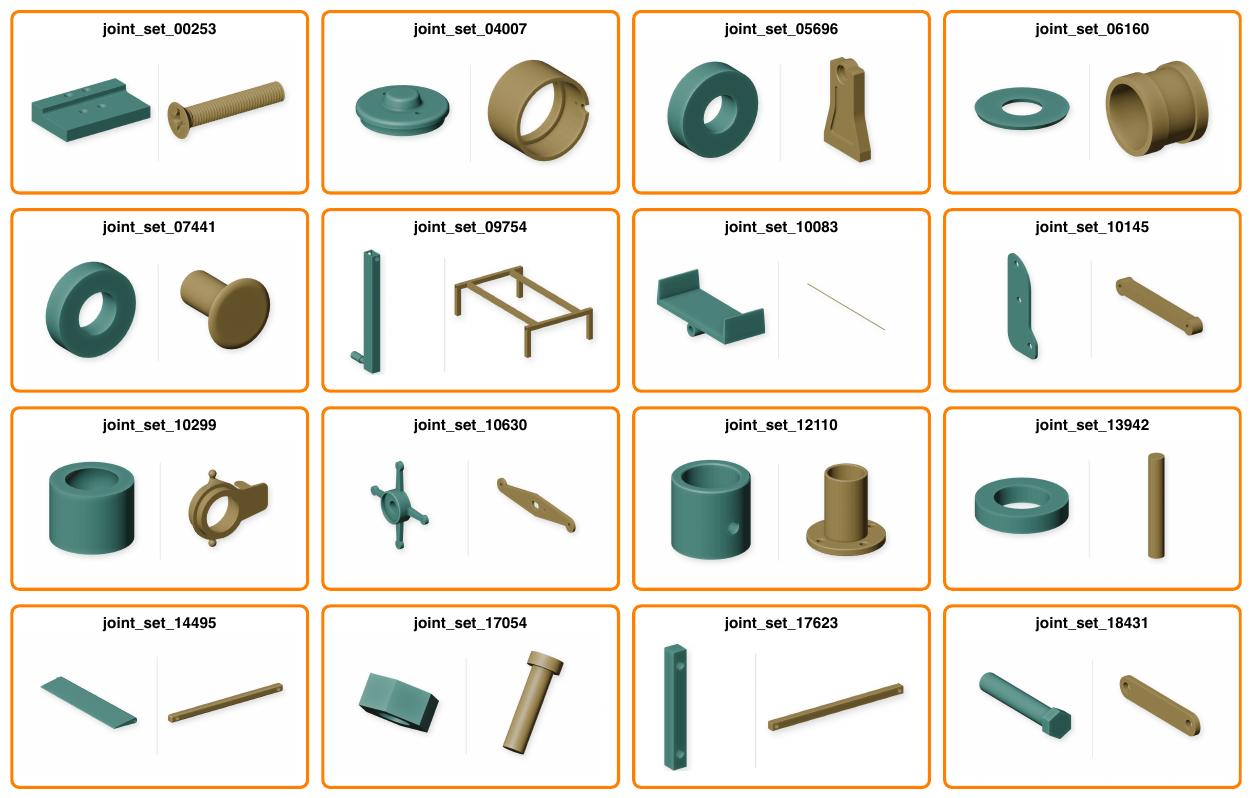}
\caption{\textbf{Additional \tracka{} body pairs.} Each cell shows the two
unassembled source bodies for one task. Joint labels, mating entities, and
frames are withheld from the model.}
\label{fig:a-gallery}
\end{figure}

\begin{table}[t]
\centering
\small
\setlength{\tabcolsep}{4.5pt}
\renewcommand{\arraystretch}{1.10}
\begin{tabular}{@{}p{0.16\textwidth}rrrp{0.44\textwidth}@{}}
\toprule
Subset & Items & Tasks & Reported items & Source evidence used for construction \\
\midrule
P-BenchCAD & 159 & 318 & 159 &
Executable CadQuery; operation history; source parameters; part family;
difficulty; standard metadata; source B-Rep \\
P-Fusion 360 & 141 & 282 & 141 &
Replayable sketch/extrude history; operation and parameter index; source
B-Rep; multiview renders \\
A-Core & 150 & 150 & 120 &
Two source bodies; B-Rep graphs; stable candidate IDs; recorded joint
alternatives; origin, axes, and motion semantics \\
\bottomrule
\end{tabular}
\caption{\textbf{Dataset composition and source evidence.} Each P part defines
independent generation and editing tasks. A results use the same 120 evaluation
pairs for every model.}
\label{tab:composition}
\end{table}

\subsection{Selection and Coverage}

P selection balances 100 easy, 100 medium, and 100 hard parts using
source structure and annotation completeness. The
resulting set covers 102 BenchCAD families, 55 CAD operations, 45
source-declared standard labels, and 12 edit types. Every part exposes one public
parameter with a measured geometric effect; 164 parts additionally support
a complete L3 analysis.

A selection is stratified by joint family and structural complexity without
using model outputs. The 150 pairs comprise 296 bodies and 17,244 indexed candidate faces or edges. They include 88 single-joint items, 46 items with
2--4 recorded joints, 13 with 5--10, and 3 with at least 11.

The complete set is covered by seven joint families: rigid, revolute, slider, cylindrical, pin-slot, planar, and ball 
relations. Family counts may overlap because one pair can contain several
recorded alternatives. The source records provide positive joint alternatives rather than an
exhaustive catalogue of every mechanically feasible relation. The absence of a
relation is not treated as a verified negative example.

\subsection{Annotation and Evidence Binding}

An engineering LLM proposes part identity, function, interfaces,
material/process context, loading context, and a functional edit. A visual VLM
independently describes global form, visible features, affordances, and
informative views. These annotations help construct the dataset's task descriptions but are not reference answers. 
A proposed annotation: dimension, feature, or edit is retained for
scoring only when source metadata, deterministic replay, or a B-Rep
measurement provides an executable check.

Because the source CAD does not provide verified operating conditions, L3 uses explicitly benchmark-assigned
analysis cases.  Each case supplies numeric material properties, support and loading regions, and limits that are applied
identically to the reference and generated geometry. These cases support controlled comparison and are not claims
about the part’s original service conditions.

\begin{table}[H]
\centering
\small
\setlength{\tabcolsep}{4pt}
\renewcommand{\arraystretch}{1.10}
\begin{tabular}{@{}p{0.18\textwidth}p{0.34\textwidth}p{0.39\textwidth}@{}}
\toprule
Task information & Evidence binding & Resulting benchmark use \\
\midrule
Part identity, function, and interfaces &
LLM/VLM proposals checked against source metadata, views, feature counts, and
B-Rep measurements &
Public brief and views; measurable dimensions and features become L1 or
preservation checks \\
Named parameter and valid range &
Source CadQuery binding or replayable Fusion history, followed by test rebuilds &
Public parameter manifest; L2-Z checks the measured parameter-to-geometry
relation at every state \\
Functional edit &
Model-assisted proposal bound to a native parameter or operation pointer &
Public edit request; L2-E checks a target measurement and protected
non-target properties \\
Material and manufacturing context &
Source label when available, otherwise a non-authoritative model-assisted
description constrained by measurable geometry &
Task context and applicable geometry-based L1/DFM checks; no manufacturing
certification is inferred \\
Physical analysis case &
Benchmark-assigned numerical material, regions, loading, and limits &
Disclosed L3 setup applied identically to reference and candidate geometry \\
Assembly relation &
Source-recorded joint converted to role-aware candidate entity pairs and a
joint frame &
Hidden A1 entity/type references and A2 frame and motion checks \\
Standards label &
Copied from source metadata &
Provenance and task context only; not independent standards certification \\
\bottomrule
\end{tabular}
\caption{\textbf{Task information and evidence binding.} Model-assisted
proposals affect scoring only after they are connected to source evidence or a
defined numerical analysis.}
\label{tab:annotation-authority}
\end{table}

\paragraph{Public and scoring data.}
Each released task contains the evidence needed to produce a response:
briefs, views, named parameters, source CAD for editing, or indexed assembly
entities as applicable. Source relations, target measurements, protected
properties, and executable predicates remain scoring data. Section
\ref{sec:prompts} specifies the exact model-facing requests; Sections
\ref{sec:p-evaluators}--\ref{sec:a-evaluators} define how the withheld references are
used.

\FloatBarrier
\section{Prompt and Output Specifications}
\label{sec:prompts}

This section documents the instructions, task inputs, and required output
formats used throughout the benchmark. For each task, the model receives a
fixed system prompt together with item-specific evidence, such as an
engineering brief, CAD views, named parameters, or indexed B-Rep entities.
The following examples show how this evidence is presented and how model
responses must be structured for executable evaluation.

\subsection{Annotation Prompts}

\begin{cebcard}[cebcarddark]{System Prompt: Engineering Scenario Annotator}
\begin{lstlisting}[style=cebprompt]
### Role and output
You are the compact engineering-scenario author for CADEngBench-P.
Return one JSON record. Every field is consumed by a benchmark prompt or a deterministic manufacturing or physics interface.

### Scenario construction
Use the supplied geometry, source labels, and fixed edit target to author one plausible benchmark scenario. State a recognizable part, application, function, one or two physical interfaces, a material and manufacturing route, one compact operating load case, and why the fixed edit matters. This is benchmark-authored engineering context, not a claim about the source author's intent.

### Material and process
Material and process: prefer source-explicit values. Otherwise select a plausible family and a short conventional specification or process variant. Use `unspecified` when a defensible choice is unavailable. Never claim compliance, test success, or source intent.

### Standards
Standards: copy supplied standard IDs when relevant. You may propose at most one additional widely recognized candidate, but never invent clauses, dimensions, ratings, or a compliance verdict. An empty list is valid.

### Load case
Load case: provide one conservative numerical design condition with the required unit pairing: force/N, torque/N_mm, pressure/MPa, or temperature/degC. Name concise application and constraint regions. Use `not_applicable`, null magnitude, and not-applicable enums only when no physical condition is defensible. The value is a benchmark scenario assumption, not verified service performance.

### Style and field rules
Write short professional phrases. Do not use Markdown, labels, paths, source IDs, placeholders, narratives, explanations, alternatives, confidence prose, or self-check text. Do not repeat the fixed edit value, parameter range, measurement method, relations, or acceptance criteria; deterministic code supplies them. Use only schema fields and keep every text value under 18 words.

- Write the design goal and engineering objective as complete engineering outcomes of at least four words.
- Keep all dimensions, magnitudes, and units exclusively in `load_case`; qualitative fields must contain no digits.
- Keep standard identifiers exclusively in `standards`. Never put a standard, size, class, schedule, rating, or compliance claim in the goal, operating condition, objective, interface, freedom, or non-requirement fields.
- Write design freedom and non-requirements as ordinary language, never parameter names, feature IDs, hashes, underscores, or colon-delimited source tokens.
- Use one material specification and one process variant, or `unspecified`; never provide alternatives such as "grade A or grade B".
- Format standard IDs conventionally with spaces and hyphens, for example `ISO 2768-1`, not underscore-separated identifiers.

### Retry corrections
Inputs may contain `retry_correction` after a rejected candidate. Follow every listed correction while independently rewriting the full record. Do not mention the rejection or copy the correction text into the output.

### Examples
Compact style example, not values to copy: mounting flange; machine guard interface; aluminum alloy; three-axis milling; static force applied at the guard interface and constrained at mounting holes; improve fastener clearance while retaining location.

- Reject: goal `torque transmission`. Prefer: `improve reliable torque transfer during manual actuation`.
- Reject: goal `maintain ASME B16.5 Class 150 compliance`. Prefer: `preserve compatible pressure-fitting interface geometry`.
- Reject: design freedom `cylinder_arg_0`. Prefer: `retain all non-target interface geometry`.
\end{lstlisting}
\end{cebcard}

\begin{cebcard}[cebcarddark]{System Prompt: Visual Geometry and Affordance Annotator}
\begin{lstlisting}[style=cebprompt]
### Role and output
You are the compact visual geometry and affordance grounder for CADEngBench-P.
Inspect only the supplied renders and public geometry facts.
Return one JSON object used directly in the zero-to-CAD prompt and visual-asset selection.

### Geometry and affordance evidence
Describe the global form, at most six reconstructive features, and at most three visually supported physical affordances. An affordance is a visible geometric capability such as mounting, locating, gripping, sealing, transmitting motion or load, or protecting another component. It is a visual hypothesis, not source intent.

### Evidence limits
Ground every feature and affordance to one or two supplied view IDs. Encode evidence quality only with `visible`, `partial`, or `occluded`. Do not infer material grade, manufacturing process, operating load, standards compliance, hidden geometry, or exact dimensions from appearance.

### Output style
Use short professional phrases. JSON field values must not contain Markdown, headings, paths, fact IDs, detector prose, uncertainty prose, alternatives, summaries, or self-check text. Put view IDs only in `view_ids`. Use only schema fields and keep every description under 18 words.

### Example
Compact style example, not values to copy: thin plate with raised boss; four openings around the boss; mounting affordance through the opening pattern; visible in both isometric views.
\end{lstlisting}
\end{cebcard}

\subsection{P Zero-to-CAD Requests}

\begin{cebcard}[cebcarddark]{System Prompt: Parametric Zero-to-CAD Generation}
\begin{lstlisting}[style=cebprompt]
### Role
You are an engineering CAD co-engineer working in a sandboxed Python environment.

### Task and evidence
Build the requested single-part model as executable CadQuery Python. The user prompt is the complete public task specification. Treat the textual requirements as authoritative for numeric dimensions, declared functional relationships, material/process context, and deliverables. When provided, the two canonical isometric renders clarify global form and feature arrangement; they do not reveal an intended construction history.

### Construction constraints
Work from the engineering outcome, not from a presumed source model. Use any valid parametric construction strategy that satisfies the stated requirements. Do not use network access, reference CAD, meshes, STEP/B-Rep input, source histories, hidden benchmark artifacts, or unavailable files. Do not claim unstated standards compliance, load ratings, material grades, or process capability.
\end{lstlisting}
\begin{lstlisting}[style=cebprompt]
### Required parametric interface
Write a complete submission.py that follows the required CadQuery interface below. DEFAULT_PARAMS contains independently editable public values. PARAMETER_MANIFEST maps each public parameter ID to the code parameter that implements it. Where the task states equality, symmetry, or repeated-feature requirements, implement the relationship in code through shared named parameters or derived values. Do not invent additional hidden constraints. build(params=None) must return the final CadQuery workplane.

### Required source structure
Follow this source structure exactly. The user prompt supplies the literal DEFAULT_PARAMS and PARAMETER_MANIFEST entries; reproduce those entries without adding construction-only values. Define build exactly once with the signature below, assign result exactly once at top level, and perform exactly one STEP export at top level after result is assigned:

\end{lstlisting}
\begin{lstlisting}[style=cebinterfacecode,language=Python]

### Required CadQuery program structure
import cadquery as cq

DEFAULT_PARAMS = {
    # Exact entries supplied by the user prompt.
}

PARAMETER_MANIFEST = {
    # Exact entries supplied by the user prompt.
}

def build(params=None):
    p = dict(DEFAULT_PARAMS)
    if params:
        p.update(params)
    # Construct the requested model using p values.
    return model

result = build()
cq.exporters.export(result, "model.step")
\end{lstlisting}
\begin{lstlisting}[style=cebprompt]

### Sandbox constraints
Do not export from inside build. Do not call globals, locals, eval, exec, open, file readers, network APIs, subprocesses, or dynamic import functions. Permitted import roots are cadquery, math, typing, collections, itertools, and functools. Avoid bare except handlers and do not catch BaseException, SystemExit, KeyboardInterrupt, or GeneratorExit.

### Output
Return only the requested program artifact. Do not include a verbal explanation, chain-of-thought, or a reconstruction of any unavailable construction history.
\end{lstlisting}
\end{cebcard}

\subsection{Representative Zero-to-CAD User Requests}

\begin{cebcard}[cebteal]{BenchCAD Request: Parallel Key}
\begin{lstlisting}[style=cebpromptcompact]
### Design request
Design an editable parametric CAD model of a parallel key for a transmission shaft interface.

- Goal: increase shaft-diameter clearance.
- Required form: elongated rectangular prism with chamfered ends and a longitudinal groove.
- Output: one solid within a nominal envelope of 23.55 x 2.36 x 2.36 mm.

### Visible geometry
- Chamfered rectangular body visible in `orbit_000` and `orbit_045`.
- Longitudinal top groove visible in `orbit_045` and `orbit_135`.
- Flat bottom and side faces with a uniform section along the length.

### Engineering context
- Function: transmit torque between a shaft and hub.
- Interface: fit within a shaft keyway.
- Affordances: locating through the groove; mounting through flat faces; load transfer through the prismatic body.
- Material and process: unspecified.
- DFM priorities: minimum wall thickness and minimum hole geometry.

### Parametric requirements
- Design freedom: clearance increase only.
- Preserve torque capacity.
Use named parameters for the declared dimensions and preserve the supplied deterministic relations.
\end{lstlisting}
\end{cebcard}

\begin{cebcard}[ceborange]{Fusion Request: Cylindrical Support}
\begin{lstlisting}[style=cebpromptcompact]
### Design request
Design an editable parametric CAD model of a cylindrical support column for an industrial machinery base.

- Goal: increase axial load capacity.
- Required form: solid right circular cylinder with uniform diameter and flat ends.
- Output: one solid within a nominal envelope of 30.0 x 300.0 x 30.0 mm.

### Visible geometry
- Continuous cylindrical outer surface visible in `orbit_045` and `orbit_270`.
- Flat circular end faces visible in `orbit_090` and `orbit_270`.

### Engineering context
- Function: transfer vertical compressive load from the machine to its foundation.
- Interface: base-plate mounting surface.
- Affordances: cylindrical press-fit or bearing surface and flat axial locating face.
- Design condition: 2.5 MPa axial pressure with the base interface constrained under static loading.
- Material and process: unspecified.
- DFM priorities: minimum wall thickness and minimum hole geometry.

### Parametric requirements
- Preserve the base footprint and hole locations.
- Surface finish is not evaluated.
Use named parameters for the declared dimensions and preserve the supplied deterministic relations.
\end{lstlisting}
\end{cebcard}

\subsection{P Functional-Edit Requests}

\begin{cebcard}[cebcarddark]{System Prompt: BenchCAD Edit}
\begin{lstlisting}[style=cebpromptcompact]
### Role
You are an AI CAD co-engineer completing a source-native engineering edit.

### Edit requirements
- Use only the supplied task evidence.
- Implement the requested change while preserving all non-target construction data and public invariants.
- Do not redesign, simplify, or approximate the part.
- Preserve the original source except for the smallest source-native edit needed to satisfy the instruction.

### Evaluation and output
The benchmark evaluates the returned artifact deterministically and does not use an LLM judge.
Return only the complete edited executable Python source, with no Markdown fences or prose.
\end{lstlisting}
\end{cebcard}

\begin{cebcard}[cebcarddark]{System Prompt: Fusion-History Edit}
\begin{lstlisting}[style=cebpromptcompact]
### Role
You are an AI CAD co-engineer completing a source-native engineering edit.

### Edit requirements
- Use only the supplied task evidence.
- Implement the requested change while preserving all non-target construction data and public invariants.
- Do not redesign, simplify, or approximate the part.
- Identify the source JSON pointer from the supplied editable-parameter index.
- Express its replacement in the declared source-native unit.
- Treat the index as exact pointers and values into the hash-bound native history, not an approximate reconstruction.

### Evaluation and output
The benchmark evaluates the returned artifact deterministically and does not use an LLM judge.
Return one JSON object matching the bounded Fusion patch schema, with no Markdown fences or prose.
\end{lstlisting}
\end{cebcard}

\begin{cebcard}[cebblue]{Edit Representations}
\small BenchCAD editing returns a complete executable Python/CadQuery program.
Fusion editing returns one bounded JSON patch against a public editable
parameter index. The five Fusion edit-operation groups reported in the main
paper describe how the patched operation combines with prior geometry; they
are not the set of primitive CAD commands.
\end{cebcard}

\begin{cebcard}[cebblue]{Deterministic Edit Scoring Paths}
\small
\textbf{BenchCAD:}\quad source CadQuery $+$ request
$\rightarrow$ complete edited CadQuery $\rightarrow$ execute and measure.\par
\vspace{3pt}
\textbf{Fusion:}\quad source history $+$ editable index $+$ request
$\rightarrow$ one bounded JSON patch $\rightarrow$ replay and measure.\par
\vspace{3pt}
Both paths require the requested target and every protected non-target
property to pass.
\end{cebcard}

\subsection{A Joint-Grounding and Kinematics Requests}

\begin{cebcard}[cebcarddark]{System Prompt: A1 Joint Grounding}
\begin{lstlisting}[style=cebassemblyprompt]
### Task
You are ranking plausible parametric joint hypotheses between two isolated CAD bodies for CADEngBench-A.

### Source relations
The source dataset consolidates alternative joints observed for identical pairs of parts across different assemblies. These are known positive configurations, not simultaneous constraints and not an exhaustive list of every mechanically possible joint. Do not predict how many source records exist.

### Available evidence
Use only the supplied standardized views, public geometry, and candidate registry. Candidate atlases color-highlight face candidates and mark edge candidates with an X at their key point. The legend prints the exact candidate ID accepted by the output schema.

### Candidate selection
- Identify likely mating features on both bodies from the clean views and highlighted atlases.
- Verify candidate IDs and analytic geometry in the registry.
- Compare geometry type, direction, radius when applicable, scale, and adjacency.
- Prefer axis-defining cylindrical faces or circular edges over incidental planar caps.
- For planar motion, prefer the functional planar face.
- Infer the permitted relative degrees of freedom before assigning the joint family.
- Use the available ranks for meaningful uncertainty in entity selection and joint family.

Return up to three distinct hypotheses in descending order of plausibility. Each hypothesis must use exactly one candidate ID from body_a and one from body_b. Rank 1 must be your best hypothesis. Never invent candidate IDs, body IDs, geometry, or hidden source facts.

### Allowed joint families
- RigidJointType: no relative translation or rotation.
- RevoluteJointType: rotation about one axis.
- SliderJointType: translation along one axis.
- CylindricalJointType: translation and rotation along/about one shared axis.
- PlanarJointType: two translations in a plane and rotation normal to it.
- BallJointType: rotation about a shared point in three rotational directions.
- PinSlotJointType: translation along a slot direction plus rotation about the pin axis.

### Output
Return one JSON object matching the supplied output schema. Do not include prose or hidden reasoning.
\end{lstlisting}
\end{cebcard}
\begin{cebcard}[cebteal]{A1 User Template}
\begin{lstlisting}[style=cebassemblyprompt]
### Assembly item
Joint set: {{assembly_id}}
Body A: {{body_a_id}}
Body B: {{body_b_id}}

### Request
Inspect the attached standardized clean views, color-highlighted candidate atlases, public geometry, and candidate registry. Rank up to three distinct plausible joint hypotheses. These hypotheses are alternatives; they are not a prediction of simultaneous constraints or source-record count.

### Output
Ground each hypothesis in the exact highlighted candidate IDs and their analytic geometry. Use the available ranks to represent genuine entity-pair uncertainty as well as joint-family uncertainty. If no defensible hypothesis can be formed, return an abstention reason. Return JSON only.
\end{lstlisting}
\end{cebcard}

\begin{cebcard}[cebcarddark]{System Prompt: A2 Joint Frame and Motion}
\begin{lstlisting}[style=cebassemblyprompt]
### Task
Predict the kinematic frame and relative placement for one selected joint hypothesis.

### Evidence and coordinates
The request supplies the same isolated CAD evidence used for A1 plus one selected ranked joint hypothesis. Use Body A's original local CAD coordinate frame and centimeters. Do not use normalized-view coordinates.

### Output constraints
Return strict JSON matching the supplied schema. Direction vectors and the rotation quaternion must be unit length. Quaternion order is [x, y, z, w]. Axis signs are physically equivalent for scoring, but the relative rotation is not. Supply only the motion-direction keys required for the selected Fusion joint family. For a rigid joint, use an empty motion_directions_body_a object and null joint origin.

### Abstention
If the geometry does not support a defensible pose, return pose: null and a structured abstention. Do not invent source limits, offsets, or hidden assembly metadata.
\end{lstlisting}
\end{cebcard}
\begin{cebcard}[ceborange]{A2 User Template}
\begin{lstlisting}[style=cebassemblyprompt]
### Selected joint
Joint set: {{assembly_id}}
Selected A1 rank: {{selected_rank}}
Selected joint type: {{joint_type}}
Body A candidate: {{body_a_candidate_id}}
Body B candidate: {{body_b_candidate_id}}

### Request
Using the attached clean views, candidate atlases, public geometry, and candidate
registry, predict:
- the joint origin in Body A local coordinates, in centimeters;
- the required motion directions in Body A local coordinates;
- the rigid transform from Body B coordinates into Body A coordinates, represented
  by translation in centimeters and quaternion [x, y, z, w].

### Output
Return JSON only.
\end{lstlisting}
\end{cebcard}
\begin{cebcard}[cebblue]{Output Contracts}
\small A1 returns up to three ranked hypotheses with joint family, Body A and
Body B candidate IDs, and confidence. A2 conditions on the selected rank-1
hypothesis and returns its joint origin, family-specific direction fields,
and the relative Body-B-to-Body-A translation and unit quaternion, all in
Body A coordinates. Exact JSON schemas are released beside the
prompts.\par\vspace{4pt}
\textbf{Scoring path.} A1 checks schema, ownership, entity pair, and
role-aware joint type. A2 validates the kinematic record, scores the joint
frame and relative placement, and authorizes PyBullet execution only after
those geometric checks pass. The simulator then tests allowed and blocked
motion.
\end{cebcard}

\FloatBarrier
\section{\trackp{} Evaluation}
\label{sec:p-evaluators}

\subsection{Inputs, Scores, and Failure Categories}

\begin{table}[t]
\centering
\small
\setlength{\tabcolsep}{4pt}
\renewcommand{\arraystretch}{1.10}
\begin{tabular}{@{}p{0.07\textwidth}p{0.22\textwidth}p{0.25\textwidth}p{0.36\textwidth}@{}}
\toprule
Layer & Evaluator input & Required evidence & Terminal failure \\
\midrule
L0 & Submitted CadQuery program and public parameters &
Process result, B-Rep solid, STEP export, independent STEP re-import &
Syntax/runtime error; timeout; no solid; invalid B-Rep; export or re-import
failure \\
L1 & L0 solid and applicable requirement cards &
B-Rep dimensions, feature/interface predicates, DFM measurements &
Any applicable engineering or DFM predicate fails \\
L2-Z & Same submitted \texttt{build(params)} and a parameter-state set &
Fresh rebuild per state; parameter-to-geometry measurement; protected
properties & Any state does not rebuild, changes the wrong feature/direction,
or violates a protected property \\
L2-E & Source CAD, one edit instruction, target measurement, preservation
checks & Edited source replay; hidden target measurement; protected
non-target measurements & Output invalid; target not achieved; or collateral
change detected \\
L3 & Generated/reference state, analysis card, material, selectors, mesh
profile, and response limits & Matched Gmsh/CalculiX outputs and derived
structural quantities & First failure among rebuild, mesh, selector, solve,
numerical validity, requirement, or FEA agreement \\
\bottomrule
\end{tabular}
\caption{\textbf{\trackp{} evaluation stages.} Each stage records the first
failure and the evidence used to determine its score.}
\label{tab:p-contract}
\end{table}

\subsection{L0 and L1: Executability and Requirements}

L0 runs the submitted program in an isolated process, calls
\texttt{build(params)}, validates every returned solid, exports STEP, and
re-imports that STEP with OpenCascade. Re-import is not a second quality
metric: it checks that the result is a portable CAD solid rather than an
in-memory object that cannot survive exchange.
This executable-artifact emphasis complements sequence- and program-level CAD
evaluation in DeepCAD, Text2CAD, and BenchCAD
\citep{wu2021deepcaddeepgenerativenetwork,
khan2024text2cadgeneratingsequentialcad,
zhang2026benchcadcomprehensiveindustrystandardbenchmark}.

L1 runs only after an L0 solid exists. Each public requirement is linked to a
deterministic measurement on the re-imported B-Rep. The current registry
checks solid count, bounding dimensions, and cylindrical-opening groups,
including count, diameter, depth, location, and axis where those quantities
are supported by the source record. A requirement without a valid binding is
reported as unscorable rather than accepted.

The DFM component contains two geometry-based screens. Minimum wall thickness
uses a certified global-thickness extractor; cylindrical-hole geometry checks
minimum diameter and maximum depth-to-diameter ratio. The benchmark limits are
1.0\,mm, 2.0\,mm, and 8, respectively. A DFM family produces pass/fail only
when applicability, a finite threshold, threshold provenance, and a certified
extractor are present. Otherwise it remains diagnostic. These checks do not
evaluate tooling access, undercuts, draft, tolerances, surface finish, CAM, or
process certification.

\begin{table}[H]
\centering
\small
\setlength{\tabcolsep}{4pt}
\renewcommand{\arraystretch}{1.10}
\begin{tabular}{@{}p{0.20\textwidth}p{0.30\textwidth}p{0.23\textwidth}
                  p{0.18\textwidth}@{}}
\toprule
L1 check & B-Rep measurement & Pass condition & Recorded failure \\
\midrule
Solid topology &
Valid-solid flag and number of solids &
Valid B-Rep with the required solid count &
Invalid topology or missing/extra body \\
Overall dimensions &
OpenCascade bounding-box lengths, with fixed axes or free axis permutation &
Every dimension lies within its source-derived tolerance &
Size or aspect mismatch \\
Cylindrical openings &
Recognized internal cylindrical faces grouped by diameter, depth, axis, and
center &
Required group count and geometric attributes match &
Missing, extra, misplaced, or incorrectly sized opening \\
Minimum wall &
Certified global minimum-thickness estimate &
$t_{\min}\geq1.0$\,mm when applicable &
Thin region or unscorable extractor \\
Hole geometry &
Minimum cylindrical-hole diameter and maximum $d_{\mathrm{depth}}/d_{\mathrm{hole}}$ &
$d_{\mathrm{hole}}\geq2.0$\,mm and ratio $\leq8$ when applicable &
Undersized/deep hole, missing required holes, or unscorable recognizer \\
\bottomrule
\end{tabular}
\caption{\textbf{L1 engineering and DFM checks.} All measurements are made on
the exported and independently re-imported B-Rep.}
\label{tab:l1-checks}
\end{table}

\begin{algorithm}[t]
\caption{L0/L1 evaluation}
\label{alg:l0l1}
\small
\begin{algorithmic}[1]
\REQUIRE submission $P$, public defaults $\theta_0$, checks $\mathcal Q_i$
\STATE Run $B\leftarrow P.\texttt{build}(\theta_0)$ in a clean process
\IF{execution fails or $B$ has no valid solid}
  \STATE \textbf{return} L0 fail
\ENDIF
\STATE Export $B$ to STEP and independently re-import it
\IF{export, re-import, or B-Rep validity fails}
  \STATE \textbf{return} L0 fail
\ENDIF
\STATE Evaluate every applicable $q(B)$ for $q\in\mathcal Q_i$
\STATE \textbf{return} L0 pass and L1 pass iff all $q(B)$ pass
\end{algorithmic}
\end{algorithm}

\subsection{L2-Z: Operational Parameterization}

L2-Z asks whether the public parameter is operational, not merely declared in
source code. The evaluator reuses one submission and rebuilds it at default,
intermediate, and valid boundary values. OpenCascade measurements check that
the named feature changes in the required direction while unrelated
properties remain invariant. This catches ignored parameters, parameters
bound to the wrong feature, and unintended coupling between dimensions.

\subsection{L2-E: Targeted Functional Editing}

L2-E is a separate task with source CAD supplied to the model. BenchCAD
submissions edit native CadQuery; Fusion submissions return one schema-bound
JSON patch that is replayed against the original construction history. A
hidden B-Rep or parameter measurement verifies the requested effect, while
preservation checks protect solid count, non-target parameters, and
item-specific geometry. Code-text similarity is not scored.

\noindent
\begin{minipage}[t]{0.485\textwidth}
\captionof{algorithm}{L2-Z parameter-family evaluation}
\label{alg:l2z}
\small
\begin{algorithmic}[1]
\REQUIRE build function $P$, values $\Theta_i$, target relation $g_i$,
protected checks $\mathcal R_i$
\FOR{each $\theta\in\Theta_i$}
  \STATE Run $B_\theta\leftarrow P.\texttt{build}(\theta)$
  \IF{$B_\theta$ is not a valid re-importable solid}
    \STATE \textbf{return} fail: rebuild
  \ENDIF
  \STATE Measure target feature and evaluate $g_i(B_\theta,\theta)$
  \STATE Evaluate every protected property $r(B_\theta)$
  \IF{target relation or preservation fails}
    \STATE \textbf{return} fail
  \ENDIF
\ENDFOR
\STATE \textbf{return} pass
\end{algorithmic}
\end{minipage}\hfill
\begin{minipage}[t]{0.485\textwidth}
\captionof{algorithm}{L2-E source-CAD edit evaluation}
\label{alg:l2e}
\small
\begin{algorithmic}[1]
\REQUIRE source CAD $S_i$, edit $E_i$, target $T_i$, checks $\mathcal R_i^E$
\STATE Apply $E_i$ through the source representation
\IF{schema, replay, solid, or STEP validation fails}
  \STATE \textbf{return} fail: invalid edit
\ENDIF
\STATE Measure target outcome $T_i(C_i^E)$
\STATE Evaluate every $r(C_i^E)$ for $r\in\mathcal R_i^E$
\IF{target or preservation fails}
  \STATE \textbf{return} fail
\ENDIF
\STATE \textbf{return} pass
\end{algorithmic}
\end{minipage}

\begin{center}
\centering
\small
\setlength{\tabcolsep}{4pt}
\begin{tabular}{@{}p{0.24\columnwidth}p{0.66\columnwidth}@{}}
\toprule
Failure evidence & Interpretation \\
\midrule
Rebuild failure & Parameter state or edit no longer creates usable CAD \\
Target relation failure & Declared parameter/edit does not control the named
feature as requested \\
Preservation failure & Non-target geometry changes as a side effect \\
Strict item pass & Every tested state or every target/protection check passes \\
\bottomrule
\end{tabular}
\captionof{table}{Interpretation of L2 failure records.}
\label{tab:l2-failures}
\end{center}


\section{Physical Verification with CalculiX}
\label{sec:l3-details}

\subsection{Analysis Specification}

CalculiX solves a supplied finite-element model. Each L3-eligible part has a
benchmark card specifying a linear-elastic material, analysis
family, deterministic support/load selectors, numeric load control,
engineering limits, and matched-response tolerances. Generated and reference
solids receive the same card at the same parameter state.

\begin{table}[H]
\centering
\small
\setlength{\tabcolsep}{3pt}
\renewcommand{\arraystretch}{1.04}
\begin{tabular}{@{}
  >{\raggedright\arraybackslash}m{0.14\textwidth}
  >{\centering\arraybackslash}m{0.25\textwidth}
  >{\raggedright\arraybackslash}m{0.23\textwidth}
  >{\raggedright\arraybackslash}m{0.28\textwidth}
@{}}
\toprule
Scenario family & Mesh and boundary conditions & Applied condition &
Compared quantities \\
\midrule
\textbf{Axial tension}\par\smallskip
$n=89$ &
\includegraphics[width=0.8\linewidth]{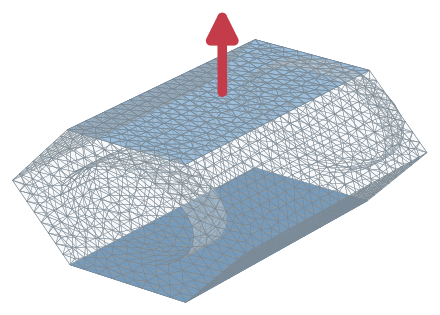} &
One end fixed; axial surface force on the opposite end &
$\sigma_{95}$, normalized maximum displacement, normalized compliance \\
\addlinespace[3pt]
\textbf{Cantilever transverse}\par\smallskip
$n=31$ &
\includegraphics[width=\linewidth]{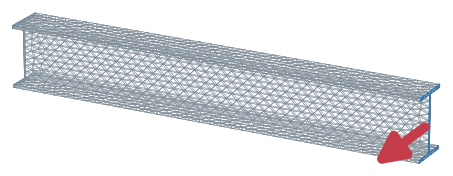} &
One end fixed; transverse surface force on the opposite end &
$\sigma_{95}$, normalized maximum displacement, normalized compliance \\
\addlinespace[3pt]
\textbf{Restrained body acceleration}\par\smallskip
$n=44$ &
\includegraphics[width=\linewidth]{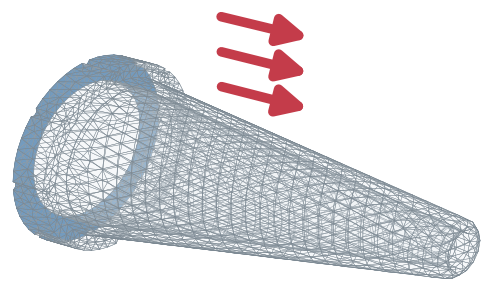} &
Mounting region fixed; acceleration applied throughout the solid &
$\sigma_{95}$, normalized maximum displacement, stress concentration \\
\bottomrule
\end{tabular}
\caption{\textbf{L3 linear-static scenario families.} Dark blue marks fixed
surfaces, light blue marks loaded surfaces, and red arrows show the applied
force or acceleration.}
\label{tab:l3-families}
\end{table}

Surface-load cases use $E=70{,}000$\,MPa, $\nu=0.33$, yield strength
250\,MPa, and safety factor 2.0. Body-acceleration cases additionally use
density $2.7\times10^{-9}$ tonne/mm$^3$. These common values isolate geometry
and response differences; they do not claim the source part was designed in
that material.

\begin{table}[H]
\centering
\scriptsize
\setlength{\tabcolsep}{4pt}
\renewcommand{\arraystretch}{1.04}
\begin{tabular}{@{}p{0.18\textwidth}p{0.34\textwidth}p{0.38\textwidth}@{}}
\toprule
Analysis-card field & Concrete example & Evaluator use \\
\midrule
Model & One solid in millimetres; linear static; small deformation &
Validates the geometry and selects the CalculiX deck and output checks \\
Material and load & Isotropic $E=70{,}000$\,MPa, $\nu=0.33$; allowable
$\sigma=125$\,MPa; fixed resultant force of 19.9\,kN &
Defines constitutive behavior, the absolute stress limit, and load magnitude \\
Boundary selectors & Opposite planar regions at the minimum and maximum
$z$ extents; normal alignment $\geq0.98$ &
Re-identifies support and load surfaces on both reference and generated solids \\
Mesh and numerics & Gmsh algorithm 10; C3D10 tetrahedra; SICN $\geq0.02$;
at most 250k elements; equilibrium error $\leq5{\times}10^{-5}$ &
Controls discretization and rejects incomplete, poor-quality, or
incorrectly loaded solves \\
Response limits & $\sigma_{95}\leq125$\,MPa; normalized displacement
$\leq0.02$; candidate/reference ratios $\leq1.25$ &
Separates solver completion from engineering-limit and matched-response passes \\
\bottomrule
\end{tabular}
\caption{\textbf{Contents of an L3 analysis card.} Values shown are from
\texttt{p\_benchcad\_000333}; the released card for each eligible part stores
the corresponding selectors, loading, mesh controls, and limits.}
\label{tab:l3-card}
\end{table}

\newpage

\subsection{Meshing and Numerical Validation}

Both solids are exported to STEP, meshed in Gmsh with quadratic tetrahedra,
and converted to CalculiX input decks \citep{geuzaine2009gmsh,
dhondt2004finiteelement}. Selector checks must identify nonempty, correctly
oriented support and load regions after every rebuild. CalculiX completion is
followed by finite-output, equilibrium, and load-control checks before any
engineering or matched-response score is computed.

\begin{algorithm}[H]
\caption{L3 matched-physics evaluation}
\label{alg:l3}
\small
\begin{algorithmic}[1]
\REQUIRE candidate/reference programs $C_i,R_i$, states $\Theta_i$,
analysis card $A_i$
\FOR{each $\theta\in\Theta_i$}
  \STATE Rebuild $C_i(\theta)$ and $R_i(\theta)$ and export STEP
  \STATE Locate support/load regions defined by $A_i$
  \IF{rebuild or selector validation fails}
    \STATE Record the first terminal failure; \textbf{continue}
  \ENDIF
  \STATE Mesh both solids with the same Gmsh profile
  \STATE Run matched CalculiX linear-static solves
  \IF{mesh, solve, finite-output, or equilibrium check fails}
    \STATE Record the first terminal failure; \textbf{continue}
  \ENDIF
  \STATE Extract stress, displacement, compliance/concentration
  \STATE Check absolute engineering limits and matched-response tolerances
\ENDFOR
\STATE Family pass iff every comparable state passes every required check
\end{algorithmic}
\end{algorithm}

\subsection{Structural Response Comparison}

Stress uses the 95th percentile of finite nodal von Mises values to reduce
single-node singularity sensitivity. Displacement is normalized by
characteristic length. Surface-load cases also compare normalized compliance;
body-acceleration cases compare a regularized stress-concentration measure.
For quantity $q$,
\begin{equation}
\delta_q(\theta)=
\left|\log\frac{q(C_i(\theta))+\epsilon_q}
{q(R_i(\theta))+\epsilon_q}\right|.
\label{eq:supp-l3}
\end{equation}
Surface-load response limits use
$\delta_q\leq\log(1.25)=0.2231$. Body-acceleration cases use 1.25 for
normalized displacement and 1.5 for stress and concentration. Absolute gates
also require stress utilization at most 1.0 and normalized displacement at
most 0.02. A solver can therefore finish and still fail either an engineering
limit or matched FEA agreement.

\begin{table}[H]
\centering
\small
\setlength{\tabcolsep}{4pt}
\begin{tabular}{@{}p{0.31\columnwidth}p{0.22\columnwidth}p{0.34\columnwidth}@{}}
\toprule
Gate & Limit & Purpose \\
\midrule
Stress utilization & $\leq1.0$ & Absolute allowable-stress screen \\
Normalized displacement & $\leq0.02$ & Absolute deformation screen \\
Surface response ratio & within $[0.8,1.25]$ & Stress, displacement,
compliance agreement \\
Body displacement ratio & within $[0.8,1.25]$ & Deformation agreement \\
Body stress/concentration & within $[2/3,1.5]$ & Stress-field agreement \\
\bottomrule
\end{tabular}
\caption{\textbf{L3 acceptance limits.} Ratio intervals are equivalent to the
symmetric log-error thresholds in the released profiles.}
\label{tab:l3-limits}
\end{table}

\FloatBarrier
\section{Assembly Joint Families}
\label{sec:joint-families}

\begin{table}[H]
\centering
\scriptsize
\setlength{\tabcolsep}{3.5pt}
\renewcommand{\arraystretch}{1.04}
\begin{tabular}{@{}m{0.08\textwidth}
                  >{\centering\arraybackslash}m{0.24\textwidth}
                  m{0.27\textwidth}m{0.31\textwidth}@{}}
\toprule
Family & Representative pair & Joint frame and allowed motion &
Executed constraints \\
\midrule
Rigid &
\includegraphics[width=0.23\textwidth,height=0.115\textheight,keepaspectratio]{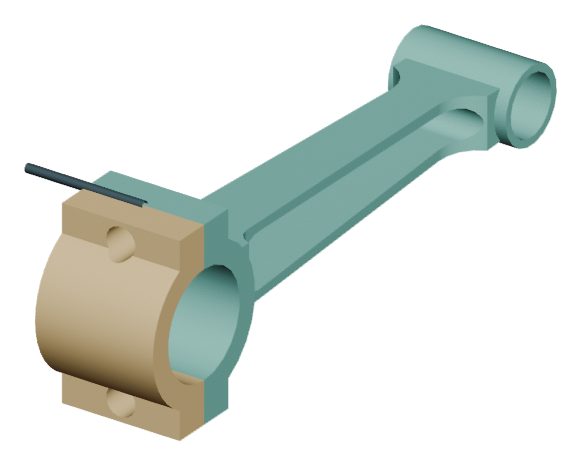} &
No motion frame; zero degrees of freedom &
Fixed joint; block all relative translation and rotation \\

Revolute &
\includegraphics[width=0.23\textwidth,height=0.115\textheight,keepaspectratio]{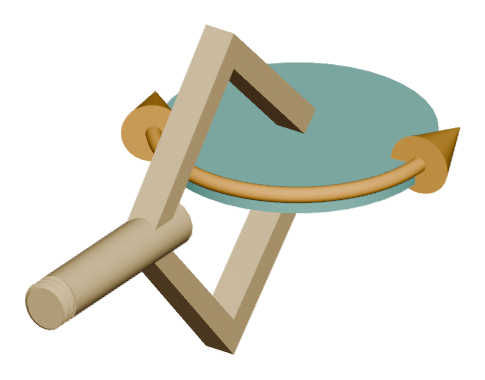} &
Origin and rotation axis; allow rotation about that axis &
Revolute joint; block translation and off-axis rotation \\

Slider &
\includegraphics[width=0.23\textwidth,height=0.115\textheight,keepaspectratio]{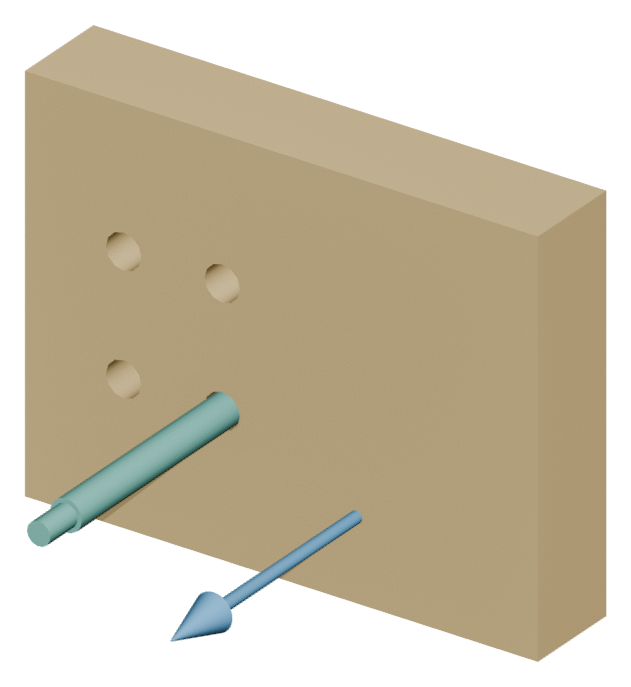} &
Origin and slide direction; allow translation along that direction &
Prismatic joint; block transverse translation and every rotation \\

Cylindrical &
\includegraphics[width=0.23\textwidth,height=0.115\textheight,keepaspectratio]{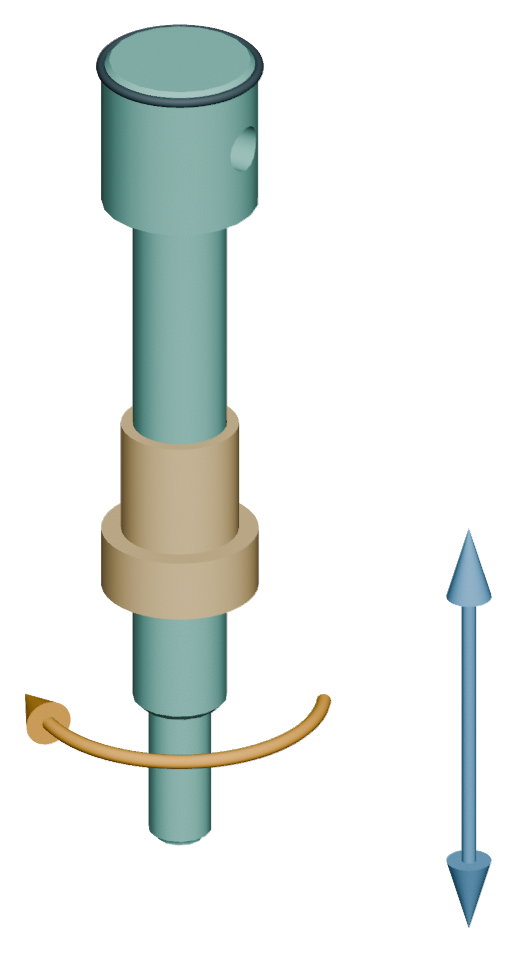} &
Origin with coaxial rotation and slide directions; allow both motions &
Revolute--prismatic compound; block transverse and off-axis motion \\

Pin-slot &
\includegraphics[width=0.23\textwidth,height=0.115\textheight,keepaspectratio]{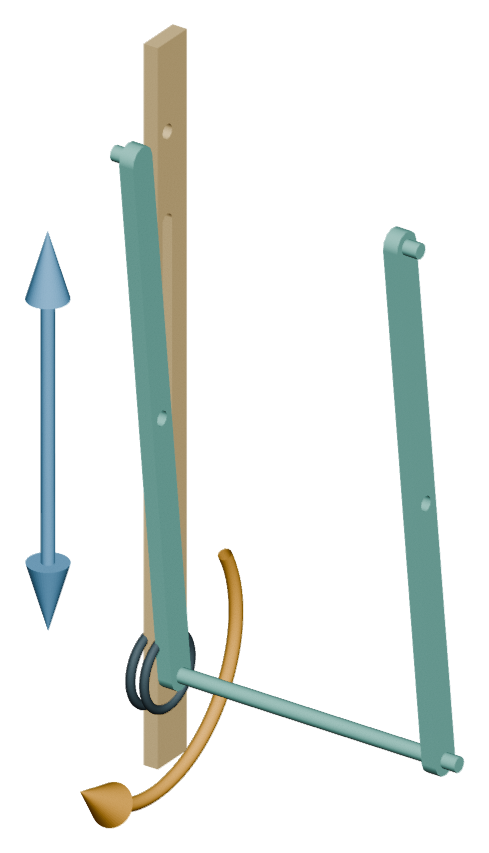} &
Origin, pin axis, and slot direction; allow pin rotation and slot translation &
Prismatic--revolute compound; block transverse translation and off-axis
rotation \\

Planar &
\includegraphics[width=0.23\textwidth,height=0.115\textheight,keepaspectratio]{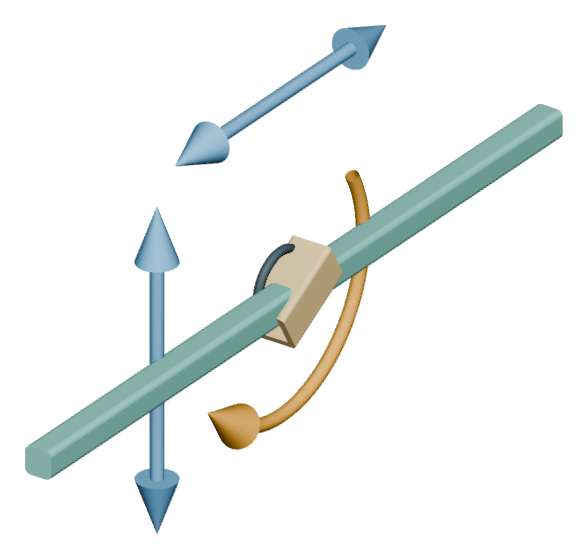} &
Origin, plane normal, and two in-plane directions; allow two translations and
normal-axis rotation &
Two-prismatic--one-revolute compound; block normal translation and
in-plane-axis rotation \\

Ball &
\includegraphics[width=0.23\textwidth,height=0.115\textheight,keepaspectratio]{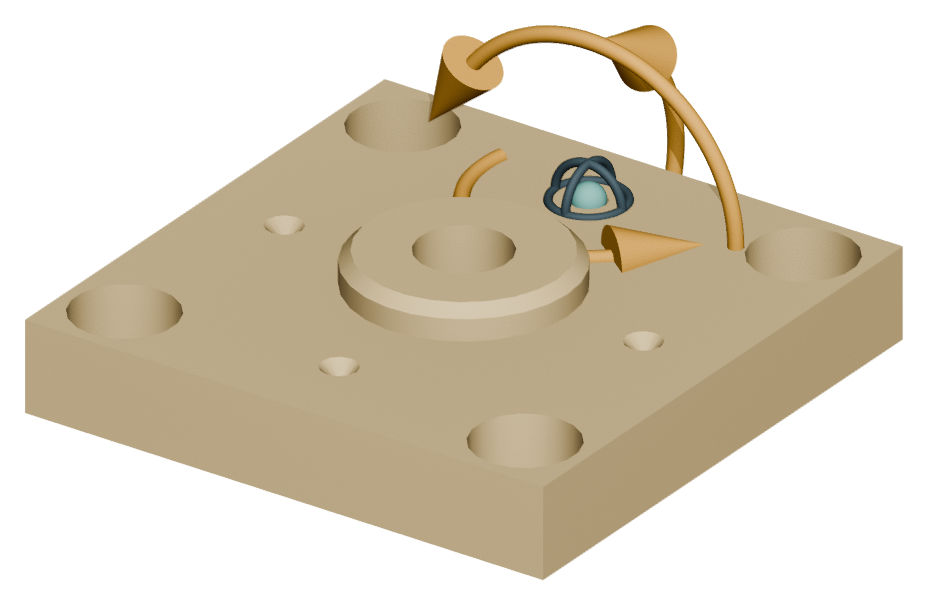} &
Origin and three orthogonal rotation directions; allow rotation about all
three axes &
Spherical joint; block translation of the joint center \\
\bottomrule
\end{tabular}
\caption{\textbf{Assembly joint families and kinematic contracts.} Each
render shows a representative source-derived body pair in its assembled pose.
Teal and gold identify the two bodies, dark outlines mark the selected B-Rep
entities, blue arrows denote translation, and orange arrows denote rotation.}
\label{tab:joint-families}
\end{table}

\FloatBarrier
\section{\tracka{} Evaluation}
\label{sec:a-evaluators}

\tracka{} evaluates an assembly relation in two stages. A1 asks the model to
rank a joint family together with one indexed B-Rep entity from each source
body. A2 then checks the frame and relative placement predicted for the
rank-1 relation and executes its allowed and blocked motion. A2 end-to-end is
reported over all 120 evaluation pairs: an A1 error is an A2 failure and does
not proceed to simulation.

\subsection{Assembly Evaluation Records}

Each task contains two unassembled source bodies from the Fusion 360 Assembly
Joint dataset. Every body is decomposed into indexed B-Rep faces and edges,
and a candidate atlas presents their stable IDs, analytic geometry types,
geometric statistics, and rendered locations. These public records let the
model refer to exact geometry without revealing the assembled pose.

The evaluator retains the source-recorded joint family, Body-A and Body-B
entities, joint frame, and relative placement. Matching is \emph{role-aware}:
an entity assigned to Body A cannot be exchanged with one from Body B. When
multiple source records identify geometrically equivalent candidates, their
verified IDs form an accepted alternative set. These are known-positive
relations rather than an exhaustive catalogue of every mechanically feasible
joint for the two bodies. The construction follows entity-grounded assembly
learning in AutoMate and JoinABLe and adds explicit frame and executed-motion
checks
\citep{jones2021automatedatasetlearningapproach,
willis2022joinablelearningbottomupassembly,
li2026assemblybenchphysicsawareassemblycomplex}.

\begin{table}[H]
\centering
\scriptsize
\setlength{\tabcolsep}{3pt}
\renewcommand{\arraystretch}{1.06}
\begin{tabular}{@{}p{0.13\textwidth}p{0.24\textwidth}p{0.27\textwidth}
                  p{0.26\textwidth}@{}}
\toprule
Artifact & Model input & Hidden reference & Evaluator use \\
\midrule
Body pair & STEP/OBJ geometry and isolated-body views & Source body identities
and assembly context & Defines the request without revealing the assembled
pose \\
B-Rep atlas & Stable candidate IDs, descriptors, and rendered locations &
Accepted role-aware and source-equivalent entity pairs & Validates IDs and
scores geometric retrieval \\
Joint record & Seven-family vocabulary & Recorded family and Body-A/Body-B
entities & Scores Entity@$k$ and Typed@$k$ \\
Frame record & Required family-specific output fields & Origin, directions,
relative translation, and relative rotation in Body-A coordinates & Validates
the predicted frame and placement \\
Motion contract & Definitions of allowed and blocked degrees of freedom &
Family-specific simulator adapter and tolerances & Executes the rank-1 joint
in PyBullet \\
\bottomrule
\end{tabular}
\caption{\textbf{Assembly evaluation records.} Public artifacts define the
model request; source-derived records provide the references used by A1 and
A2.}
\label{tab:a-construction}
\end{table}

\subsection{A1: Exact Joint and Entity Retrieval}

A1 asks the model to return up to three alternatives in descending confidence.
Each hypothesis $h_k=(j_k,e_k^A,e_k^B,s_k)$ contains a joint family, one
candidate ID from each body, and a confidence score. \textbf{Valid} requires
strict first-pass schema-valid JSON, the correct assembly ID, consecutive
ranks starting at one, recognized joint families, unique hypotheses,
confidence values in $[0,1]$, and candidate IDs that exist on the declared
body side. An abstention is valid only when the hypothesis list is empty.
Syntax-only recovery is recorded separately and does not repair semantic
errors.

\textbf{Entity@$k$} passes when an accepted Body-A/Body-B entity pair appears
within the first $k$ hypotheses, independent of its predicted family.
\textbf{Typed@$k$} additionally requires that the family and entity pair match
the same source-recorded relation. \textbf{MRR} is the reciprocal rank of the
first typed-correct hypothesis. A family-only prediction, swapped body roles,
or two individually plausible but incorrectly paired entities does not pass.

\begin{algorithm}[H]
\caption{A1 role-aware joint retrieval}
\label{alg:a1}
\small
\begin{algorithmic}[1]
\REQUIRE ranked hypotheses $H$, Body A/B atlases, accepted source relations
\STATE Validate the response schema, assembly ID, ranks, vocabulary, and IDs
\IF{the response is invalid}
  \STATE \textbf{return} Valid fail and zero retrieval credit
\ENDIF
\FOR{$k=1,\ldots,\min(3,|H|)$}
  \STATE Parse family $j_k$, entities $e_k^A,e_k^B$, confidence $s_k$
  \STATE Compare $(e_k^A,e_k^B)$ with accepted role-aware entity pairs
  \STATE Record Entity@$k$; also require $j_k$ for Typed@$k$
\ENDFOR
\STATE Compute reciprocal rank of the first fully typed-correct hypothesis
\end{algorithmic}
\end{algorithm}

\subsection{A2: Joint Frame and Executed Motion}

A2 evaluates the model's rank-1 A1 prediction without replacing it with the
reference relation. The output specifies the joint origin, family-specific
axes or directions, and the relative Body-B-to-Body-A translation and unit
quaternion. Rigid joints omit the origin and motion directions; every
non-rigid family requires an origin and the fields listed in
Table~\ref{tab:joint-families}. The evaluator first checks this geometric
frame and relative placement. A typed-correct prediction with a well-formed,
mechanically representable frame may be converted to a PyBullet joint and
tested for motion even when a frame-error tolerance is exceeded. Such a
prediction remains an A2 E2E failure.

\begin{table}[H]
\centering
\scriptsize
\setlength{\tabcolsep}{3.6pt}
\renewcommand{\arraystretch}{1.08}
\begin{tabular}{@{}p{0.16\textwidth}p{0.30\textwidth}p{0.24\textwidth}
                  p{0.20\textwidth}@{}}
\toprule
Gate & Quantity checked & Pass condition & Failure outcome \\
\midrule
A1 grounding & Rank-1 family and role-aware entity pair & Typed-correct
rank-1 hypothesis & A2 E2E fail; simulation not authorized \\
Representation & Required fields; vector and quaternion norms & Family fields
present; norm error $\leq10^{-3}$ & Invalid frame record \\
Joint origin & Euclidean origin error normalized by Body-A bounding-box
diagonal & $\leq1\%$ & Frame failure; simulation may proceed if representable \\
Relative pose & Translation error normalized by the larger body diagonal;
quaternion angular error & Translation $\leq1\%$; rotation $\leq5^\circ$ &
Placement failure \\
Directions & Sign-invariant angular error for each required axis or direction &
$\leq5^\circ$ & Frame failure \\
Allowed motion & Prescribed joint coordinate and motion-axis agreement &
Coordinate error $\leq10^{-6}$; axis dot product $>0.999999$ & Motion failure \\
Stability & Drift during a 240-step hold & Drift $<10^{-5}$ & Stability failure \\
Blocked motion & Leakage under off-axis force or torque & Direct drift
$<10^{-4}$; compound normalized leakage $<10^{-3}$ & Constraint failure \\
\bottomrule
\end{tabular}
\caption{\textbf{A2 evaluation gates.} Body-size normalization makes frame
thresholds comparable across differently scaled parts. Motion thresholds are
applied after the predicted joint is instantiated.}
\label{tab:a2-contracts}
\end{table}

For simulation, Body A is fixed and the predicted Body-B placement is used
unchanged. PyBullet runs with zero gravity, a $1/240$\,s timestep, and 240
steps. Direct revolute or prismatic joints use 100 solver iterations;
cylindrical, planar, pin-slot, and spherical adapters use 200. Allowed-motion
probes prescribe $0.25$\,rad rotation or $0.01$\,m translation, clamped inside
recorded source limits. Ball joints are tested about all three predicted axes.
Where applicable, unintended rotation must remain below $10^{-7}$.

Blocked-motion probes apply off-axis forces and torques instead of trusting
the declared joint type. Direct fixtures use 20\,N and 10\,N\,m. Compound and
spherical adapters scale the perturbation using the moving body's mass and
principal inertia, then compare constrained leakage with the corresponding
unconstrained motion. Collision contacts are retained as diagnostics but do
not determine the score by themselves.

\paragraph{End-to-end scoring.}
Every one of the 120 evaluation pairs remains in the A2 E2E denominator. An
incorrect A1 relation, malformed frame, frame or placement error, simulator
failure, or failed motion probe produces an A2 E2E failure. The number of
executed simulations therefore measures how many predictions reached the
simulation gate, not A2 accuracy; a representable frame may be simulated while
remaining an E2E failure.

\begin{algorithm}[H]
\caption{A2 conditional frame and motion evaluation}
\label{alg:a2}
\scriptsize
\begin{algorithmic}[1]
\REQUIRE rank-1 A1 hypothesis, predicted record
$F=(o,D,t_{B\rightarrow A},q_{B\rightarrow A})$, source bodies
\IF{rank-1 family/entity relation is not fully correct}
  \STATE \textbf{return} A2 E2E fail; do not authorize simulation
\ENDIF
\STATE Validate required fields, body frame, unit vectors, and unit quaternion
\STATE Compare origin, directions, relative translation, and relative rotation
\IF{the predicted joint is malformed or mechanically unrepresentable}
  \STATE \textbf{return} A2 E2E fail; do not authorize simulation
\ENDIF
\STATE Build the predicted joint in Body A coordinates
\STATE Execute allowed-motion, stability, and blocked-motion probes
\STATE \textbf{return} pass iff frame thresholds and all motion probes pass
\end{algorithmic}
\end{algorithm}

\FloatBarrier
\section{Benchmarking Protocol}
\label{sec:benchmark-protocol}

Within each track, every system receives the same public item evidence, system
instructions, and output schema. Evaluation is one-shot: each model produces
one response for each task, with no semantic retry or manual repair.
Malformed responses, timeouts, and build failures are retained as failures.

\trackp{} reports results over all 300 parts. For \tracka{}, 30 of the 150
body pairs were used to test prompts, parsers, and evaluator execution before
the model comparison. These pairs were then excluded; every reported
\tracka{} result uses the same remaining 120 pairs.

\begin{table}[H]
\centering
\scriptsize
\setlength{\tabcolsep}{2.8pt}
\renewcommand{\arraystretch}{1.08}
\begin{tabular}{@{}p{0.14\textwidth}p{0.23\textwidth}p{0.19\textwidth}
                  p{0.09\textwidth}p{0.07\textwidth}p{0.08\textwidth}
                  p{0.10\textwidth}@{}}
\toprule
Reported name & Model identifier & Access route & Reasoning &
Temp. & Top-$p$ & P output cap \\
\midrule
GPT-5.2 & \path{gpt-5.2-2025-12-11} &
OpenAI API & None & 0 & 1.0 & 16,384 \\
Claude 4.5 & \path{claude-sonnet-4-5-20250929} &
Anthropic API via LiteLLM & None & 0 & default & 16,384 \\
Gemini 3 Flash & \path{gemini-3-flash-preview} &
Google API & Minimal & 1.0 & 1.0 & 16,384 \\
GLM-4.6V & \path{z-ai/glm-4.6v} &
OpenRouter / Novita & None & 0 & 1.0 & 16,384 \\
Kimi K2.5 & \path{moonshotai/kimi-k2.5} &
OpenRouter / DigitalOcean & None & 0 & 1.0 & 16,384 \\
Mistral 3.5 & \path{mistral-medium-3-5} &
Mistral API & None & 0 & 1.0 & 16,384 \\
Llama 4 Maverick & \path{meta-llama/llama-4-maverick} &
OpenRouter / DeepInfra FP8 & None & 0 & 1.0 & 16,384 \\
Qwen3.5-35B & \path{qwen/qwen3.5-35b-a3b-20260224} &
OpenRouter FP8 & None & 0.7 & 0.8 & 32,768 \\
\bottomrule
\end{tabular}
\caption{\textbf{Model versions and P generation settings.}}
\label{tab:model-configurations}
\end{table}

All scoring is performed after generation with a common evaluator for every
model. P programs execute in isolated CadQuery processes and their B-Reps are
measured with OpenCascade. L3 uses Gmsh 4.15.2, CalculiX 2.21, and quadratic
C3D10 tetrahedra. For A2, a valid rank-1 joint prediction is instantiated on
the source bodies and its allowed and blocked motions are executed in
PyBullet.
\FloatBarrier
\section{Stage-Wise Parametric CAD Examples}
\label{sec:worked-cases}

These five observed submissions illustrate the sequential parametric-CAD
evaluation path. The cards are ordered
by the earliest failed gate: L0 establishes that an executable CAD artifact
exists; L1 checks its default geometry and engineering requirements; L2-Z
tests whether its named parameters control the intended geometry; and L3
compares its FEA behavior with the parameter-matched reference.

\begin{cebcard}[cebcoral]{L0 Execution Failure}
\begin{minipage}[c]{0.25\linewidth}
\centering
\includegraphics[width=0.92\linewidth]{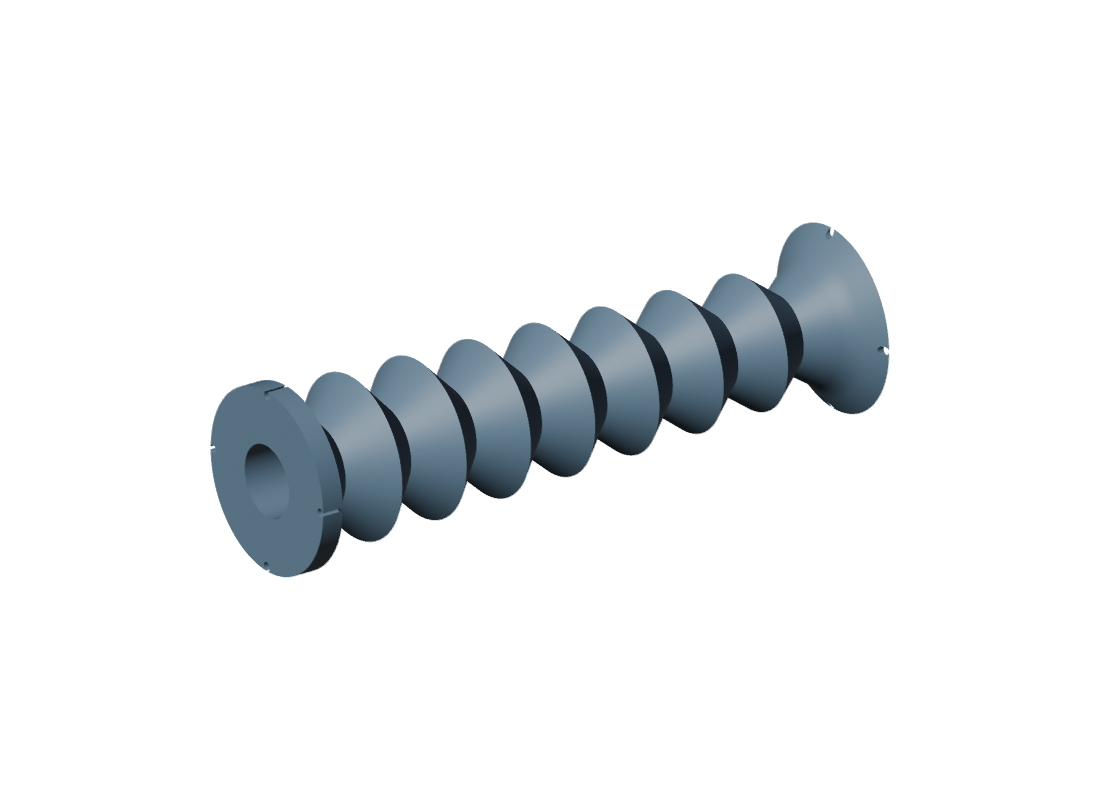}\\[-2pt]
{\scriptsize Public task geometry; no candidate solid was produced}
\end{minipage}\hfill
\begin{minipage}[c]{0.70\linewidth}
\textbf{Kimi K2.5} \hfill
\texttt{\scriptsize p\_benchcad\_000038}\\[3pt]
\textbf{Path:} \failmark\ L0 \quad \nrmark\ L1--L3
\medskip

\begin{tabular}{@{}p{0.35\linewidth}p{0.66\linewidth}@{}}
Static program contract & \passmark \\
Fresh-process execution & \failmark:
\texttt{\scriptsize BRep\_API: command not done} \\
STEP and B-Rep checks & Not reached \\
\end{tabular}
\medskip

The response has the required program interface, but its CadQuery construction
raises an OpenCascade error. No geometry exists for requirements, parameter
replay, or FEA.
\end{minipage}
\end{cebcard}

\begin{cebcard}[cebcoral]{L1 Requirement Failure}
\begin{minipage}[c]{0.25\linewidth}
\centering
\includegraphics[width=0.95\linewidth]{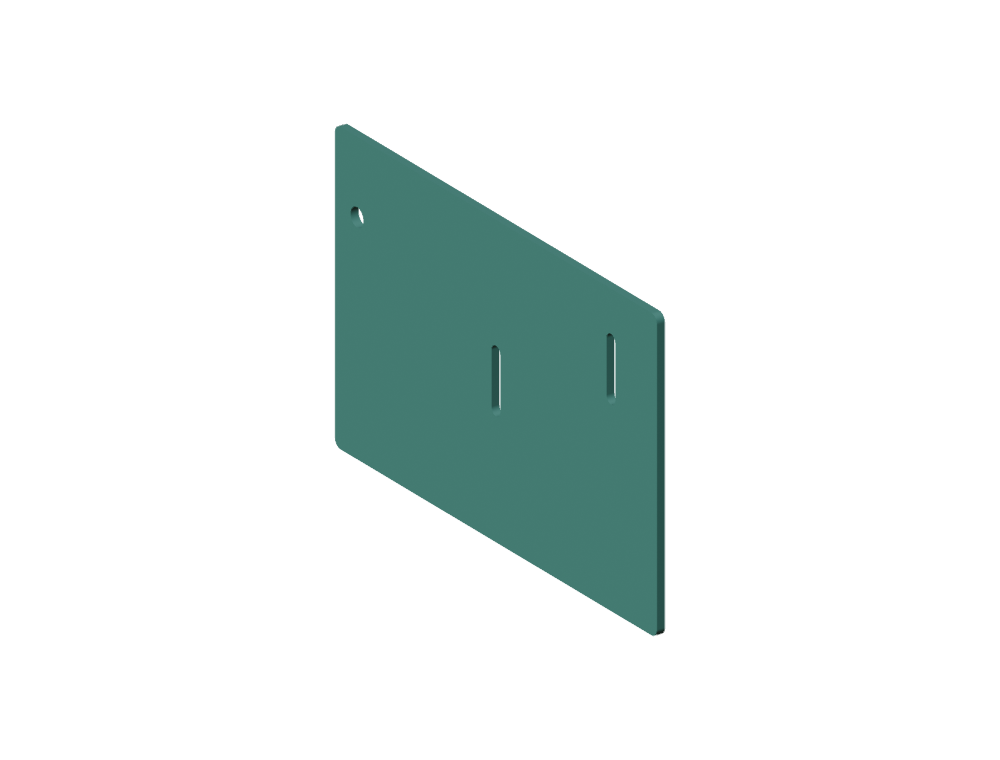}\\[-2pt]
{\scriptsize Llama 4 Maverick candidate}
\end{minipage}\hfill
\begin{minipage}[c]{0.70\linewidth}
\textbf{Llama 4 Maverick} \hfill
\texttt{\scriptsize p\_benchcad\_000068}\\[3pt]
\textbf{Path:} \passmark\ L0 \quad \failmark\ L1 (first) \quad
\failmark\ L2-Z \quad \nrmark\ L3
\medskip

\begin{minipage}[c]{0.46\linewidth}
L0 exports one valid STEP solid with the expected global envelope. L1 then
recognizes none of four required 5.42\,mm openings and only one of two required
11.56\,mm openings. The DFM feature-count check therefore observes one of six
required cylindrical openings.
\end{minipage}\hfill
\begin{minipage}[c]{0.51\linewidth}
\centering
\includegraphics[width=\linewidth]{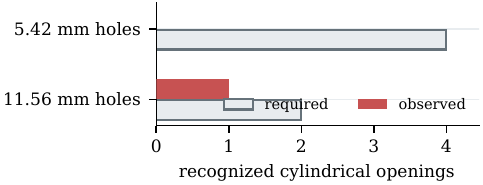}
\end{minipage}
\end{minipage}
\end{cebcard}

\begin{cebcard}[ceborange]{L2-Z Parameter-Response Failure}
\begin{minipage}[c]{0.25\linewidth}
\centering
\includegraphics[width=0.95\linewidth]{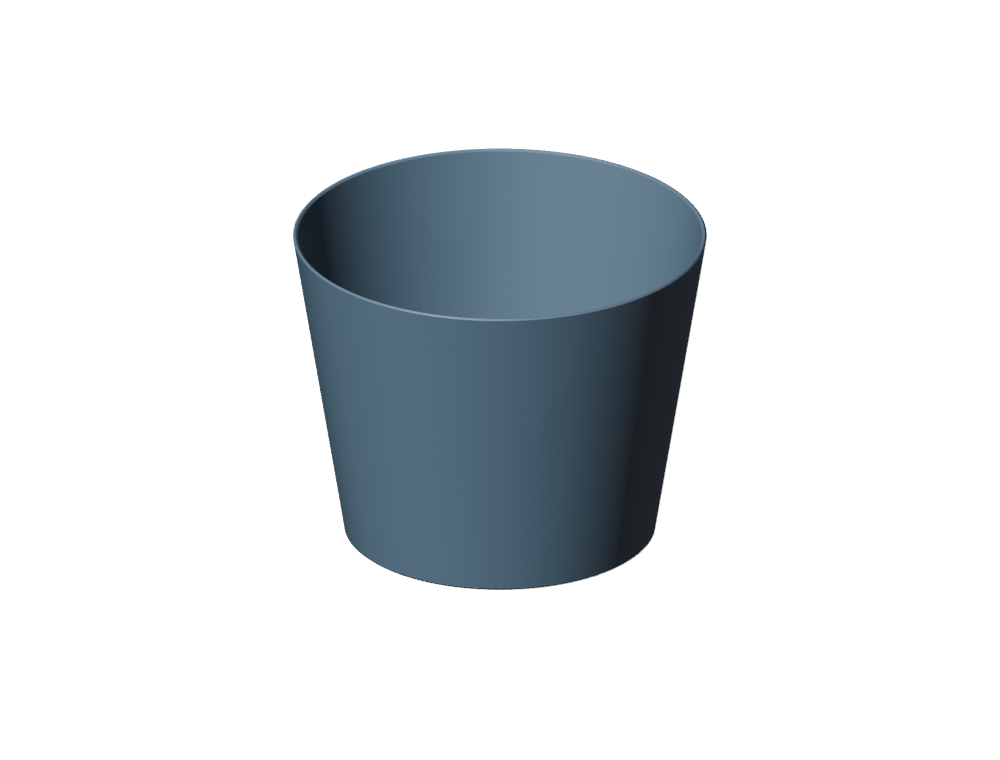}\\[-2pt]
{\scriptsize Gemini 3 Flash candidate}
\end{minipage}\hfill
\begin{minipage}[c]{0.70\linewidth}
\textbf{Gemini 3 Flash} \hfill
\texttt{\scriptsize p\_benchcad\_000062}\\[3pt]
\textbf{Path:} \passmark\ L0--L1 \quad \failmark\ L2-Z \quad \nrmark\ L3
\medskip

\begin{minipage}[c]{0.46\linewidth}
The default solid executes and satisfies its L1 checks. L2-Z then replays six
values of \texttt{chamfer\_arg\_0}. Every build returns the same geometry
fingerprint and volume, so the declared parameter does not control the
requested chamfer.
\end{minipage}\hfill
\begin{minipage}[c]{0.51\linewidth}
\centering
\includegraphics[width=\linewidth]{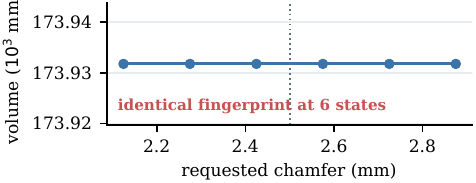}
\end{minipage}
\end{minipage}
\end{cebcard}

\begin{cebcard}[cebcoral]{L3 FEA Comparison Failure}
\begin{minipage}[c]{0.25\linewidth}
\centering
\includegraphics[width=0.95\linewidth]{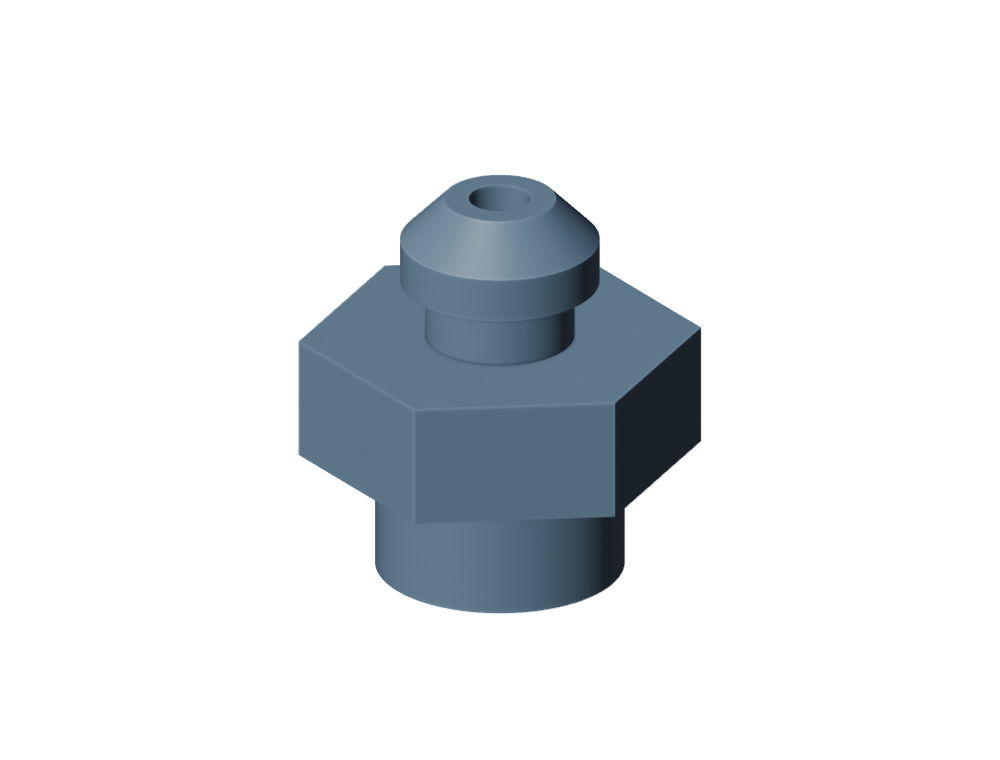}\\[-2pt]
{\scriptsize Gemini 3 Flash candidate}
\end{minipage}\hfill
\begin{minipage}[c]{0.70\linewidth}
\textbf{Gemini 3 Flash} \hfill
\texttt{\scriptsize p\_benchcad\_000258}\\[3pt]
\textbf{Path:} \passmark\ L0--L2-Z \quad \failmark\ L3 (0/3)
\medskip

\begin{minipage}[c]{0.46\linewidth}
All three parameter states rebuild, mesh, and solve under the matched axial
tension case. The candidate nevertheless produces only about 61\% of the
reference stress and 51\% of its normalized displacement. Solver completion
alone therefore does not satisfy L3.
\end{minipage}\hfill
\begin{minipage}[c]{0.51\linewidth}
\centering
\includegraphics[width=\linewidth]{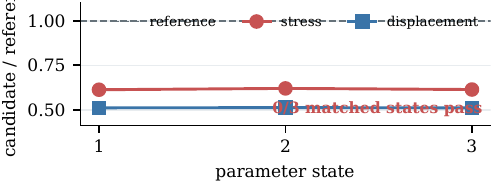}
\end{minipage}
\end{minipage}
\end{cebcard}

\begin{cebcard}[cebteal]{Complete L0--L3 Pass}
\begin{minipage}[c]{0.25\linewidth}
\centering
\includegraphics[width=0.95\linewidth]{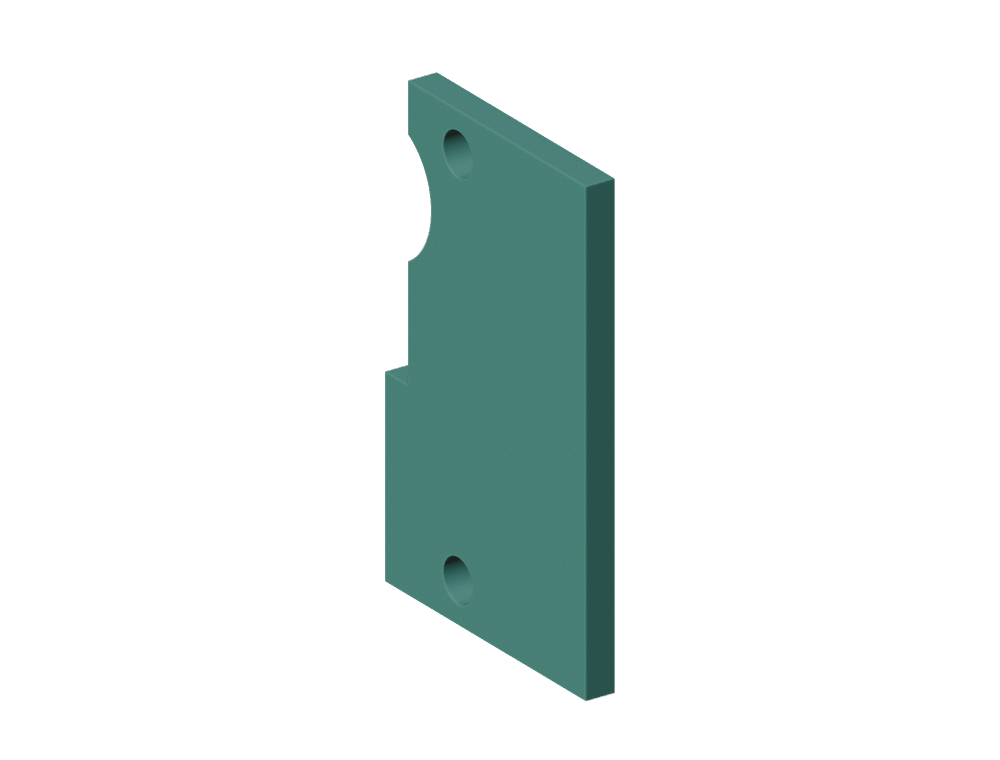}\\[-2pt]
{\scriptsize GPT-5.2 candidate}
\end{minipage}\hfill
\begin{minipage}[c]{0.70\linewidth}
\textbf{GPT-5.2} \hfill
\texttt{\scriptsize p\_benchcad\_000393}\\[3pt]
\textbf{Path:} \passmark\ L0--L3 (7/7)
\medskip

\begin{minipage}[c]{0.46\linewidth}
The program exports a valid solid, satisfies the default requirements, and
changes the named dimension in every replayed state. All seven matched
cantilever cases also mesh and solve, with stress and normalized displacement
remaining within the reference tolerances.
\end{minipage}\hfill
\begin{minipage}[c]{0.51\linewidth}
\centering
\includegraphics[width=\linewidth]{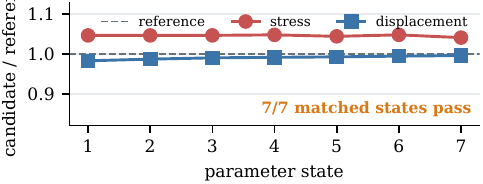}
\end{minipage}
\end{minipage}
\end{cebcard}

Successful execution does not establish that the requested
features exist; correct default geometry does not establish that the program
is genuinely parametric; and a valid parameterized family can still produce
an incorrect structural response. The benchmark therefore applies the layers
as a gated hierarchy: a submission receives credit at a stage only after
satisfying the prerequisites needed to interpret that result. This prevents
an incidental downstream success from masking an upstream failure, while
locating the earliest unsupported capability.

\section{Pin-Slot Joint Case Study}
\label{sec:pin-slot-case}

Assembly performance varies sharply by relation family. Mean Typed@3 is
59.9\% on 29 revolute items and 49.4\% on 21 cylindrical items, but only
10.0\% on planar and 0/6 on pin-slot for every tested model. These slices
overlap when a pair records multiple families. The shared pin-slot zero is a
useful diagnostic observation, but six items are too few to support a broad
claim about all pin-slot assemblies.

\begin{center}
\centering
\includegraphics[width=\textwidth]{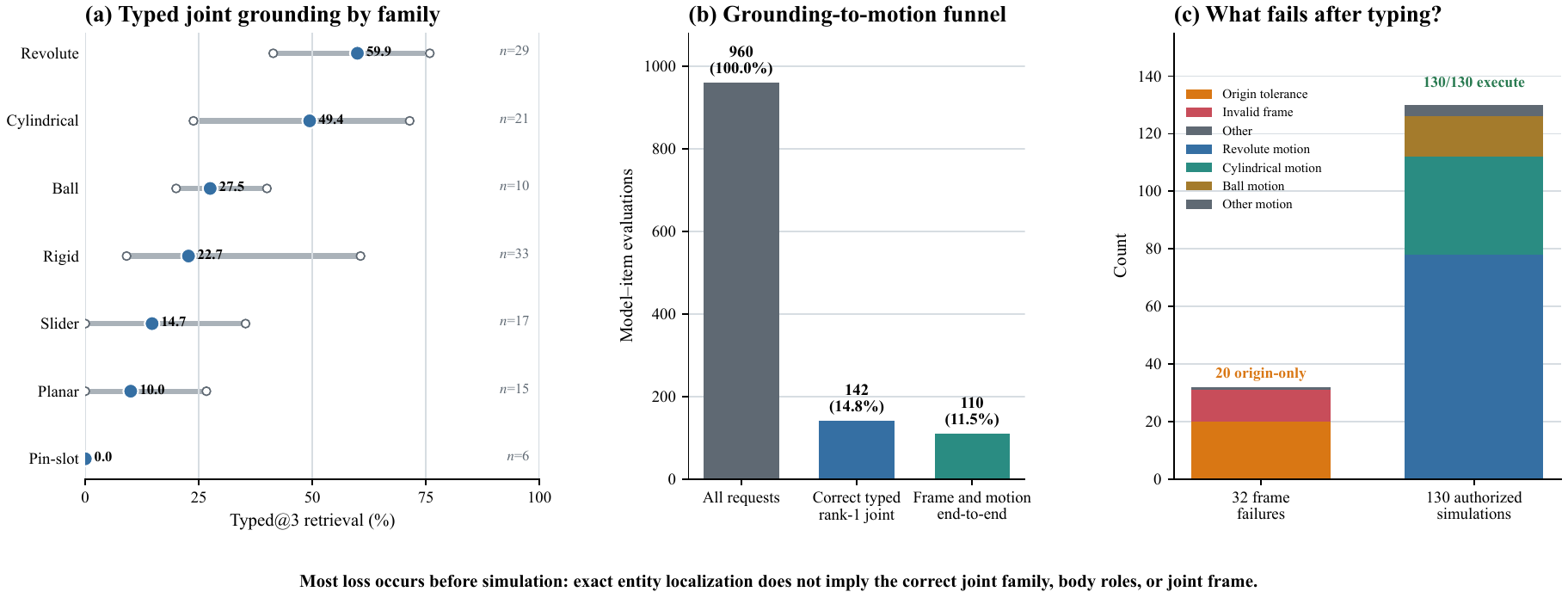}
\captionof{figure}{\textbf{From B-Rep retrieval to executed motion.}
\textbf{(a)} Dots are model means and lines are model ranges; family slices
overlap when a pair has multiple source-recorded families.
\textbf{(b--c)} Most end-to-end loss occurs before an authorized simulation.}
\label{fig:a-evidence}
\end{center}

\begin{center}
\centering
\IfFileExists{figures/supplement/pin_slot_case_visual.pdf}{%
  \includegraphics[width=\textwidth]{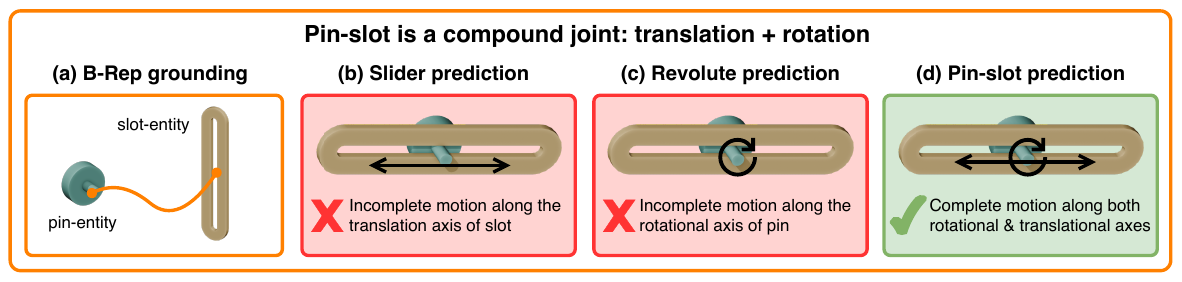}%
}{%
  \begin{tcolorbox}[
    enhanced,
    colback=ceblight,
    colframe=cebcardline,
    boxrule=0.55pt,
    arc=2.5pt,
    outer arc=2.5pt,
    left=8pt,
    right=8pt,
    top=12pt,
    bottom=12pt
  ]
  \centering
  \textbf{Pin-slot visual reserved: \texttt{joint\_set\_17452}}\\[5pt]
  \small Source bodies $\rightarrow$ exact slot/pin B-Rep entities
  $\rightarrow$ assembled frame $\rightarrow$ coupled translation and
  rotation. The final visual contrasts the correct pin-slot relation with
  revolute-only and slider-only hypotheses.
  \end{tcolorbox}
}
\captionof{figure}{\textbf{Pin-slot grounding and motion.}
The joint must bind the pin and slot entities and recover both motion
components; selecting only a rotation axis or only a translation direction is
incomplete.}
\label{fig:pin-slot-visual}
\end{center}

\begin{cebcard}[ceborange]{Pin-Slot Evaluation Requirements}
\small Revolute and cylindrical interfaces often expose a dominant shared
axis. Pin-slot instead combines entity grounding with two linked motion
semantics: translation along the slot and rotation about the pin. The
aggregate results show that this slice was unresolved in the reported
evaluation; the visual identifies the geometric and kinematic information
required by a correct prediction.
\end{cebcard}

\FloatBarrier
\section{Statistical Analysis}
\label{sec:statistics}

\subsection{Confidence Intervals}

Each reported pass rate is an estimate from a finite collection of CAD tasks.
The confidence intervals below quantify how sensitive that estimate is to the
particular benchmark items included. For example, a cell reported as
$41.3\,[35.7,47.0]$ contains the observed pass rate followed by its 95\%
bootstrap interval. A narrower interval indicates a more stable estimate
across resampled task sets.

The evaluated population differs by metric. L0, L1, and L2-E use all 300 P
parts. L2-Z reports its model-specific evaluable denominator (287--300
parts). L3 pair pass includes only parameter states for which both the
generated and reference designs produce comparable FEA results, so it is
reported together with L3 reach. Every A model is evaluated on the same 120
body pairs.

For P, we draw 10,000 bootstrap samples, stratified by source dataset and
clustered by part (seed 20260726). Clustering keeps all parameter states and
checks derived from one part in the same resample. For A, we draw 2,000
samples grouped by reconstructed assembly graph, so related body pairs are
not treated as independent tasks. The resulting intervals measure uncertainty
from benchmark composition. They do not measure variation across repeated
model generations because each model produced one response per request.

\noindent
\begin{minipage}[t]{0.56\textwidth}
\centering
\tiny
\setlength{\tabcolsep}{1.6pt}
\input{tables/supplement_p_confidence_intervals.tex}
\captionof{table}{\textbf{P estimates with 95\% item-clustered bootstrap
intervals.} Cells show percent [lower, upper]. L3 is conditional pair pass.}
\label{tab:p-ci}
\end{minipage}\hfill
\begin{minipage}[t]{0.42\textwidth}
\centering
\tiny
\setlength{\tabcolsep}{1.7pt}
\input{tables/supplement_a_confidence_intervals.tex}
\captionof{table}{\textbf{A estimates with 95\% grouped-bootstrap intervals.}
Typed metrics are percentages; MRR is on $[0,1]$.}
\label{tab:a-ci}
\end{minipage}

\subsection{Paired Model Comparisons}

Confidence intervals describe uncertainty in each model's score, but they do
not directly test whether two models differ. Because every system is evaluated
on the same tasks, L0--L2 comparisons use an exact McNemar test on paired
binary outcomes. The test considers the discordant items: cases where Model A
passes and Model B fails, and cases where Model B passes and Model A fails. A
difference is supported when these two counts are sufficiently imbalanced,
rather than merely when the aggregate percentages differ.

Eight models yield 28 pairwise comparisons for each metric. We therefore
apply Holm correction separately within each metric to control false
discoveries from multiple testing. Table~\ref{tab:p-significance} reports how
many of the 28 comparisons remain significant at $\alpha=0.05$ after this
correction; it does not report the magnitude of the score differences.

No L3 pairwise comparison survives correction. L3 uses a smaller,
model-dependent set of comparable FEA states, so this result indicates
insufficient evidence for a corrected pairwise difference, not equivalent
structural behavior across models. A results use the grouped-bootstrap
intervals above. Joint-family slices remain descriptive because one body pair
may contain more than one recorded family.

\begin{center}
\centering
\small
\input{tables/supplement_p_significance_counts.tex}
\captionof{table}{\textbf{Holm-corrected paired comparisons at $\alpha=0.05$.} Each
row contains all $\binom{8}{2}=28$ model pairs.}
\label{tab:p-significance}
\end{center}

\begingroup
\small
\setlength{\bibsep}{1pt}
\renewcommand{\refname}{Supplementary References}
\bibliography{references}
\endgroup

%% file: tables/supplement_p_confidence_intervals.tex
\begin{tabular}{@{}lccccc@{}}
\toprule
System & L0 & L1 & L2-Z & L2-E & L3 pair pass \\
\midrule
GPT-5.2 & 41.3 [35.7, 47.0] & 22.0 [17.3, 26.7] & 31.6 [26.3, 37.2] & 70.3 [65.3, 75.0] & 32.8 [20.2, 46.0] \\
Claude 4.5 & 58.0 [52.3, 63.3] & 28.3 [23.3, 33.7] & 41.4 [35.7, 46.9] & 66.7 [61.3, 71.7] & 35.8 [25.3, 46.7] \\
Gemini 3 Flash & 52.7 [47.0, 58.3] & 30.0 [25.0, 35.3] & 39.8 [34.3, 45.6] & 72.3 [67.0, 77.3] & 46.3 [34.4, 58.0] \\
GLM-4.6V & 47.3 [41.7, 53.0] & 13.3 [9.7, 17.3] & 30.7 [25.4, 36.1] & 59.3 [54.0, 64.3] & 27.0 [16.3, 38.4] \\
Kimi K2.5 & 53.3 [47.7, 59.0] & 24.3 [19.7, 29.3] & 41.0 [35.2, 46.7] & 67.0 [61.7, 72.0] & 34.8 [24.0, 45.8] \\
Mistral 3.5 & 29.3 [24.3, 34.7] & 8.3 [5.3, 11.7] & 16.5 [12.4, 20.9] & 68.3 [63.0, 73.3] & 34.5 [20.5, 48.9] \\
Llama 4 Maverick & 48.0 [42.7, 53.7] & 12.7 [9.0, 16.3] & 34.0 [28.6, 39.5] & 60.7 [55.3, 66.0] & 27.5 [16.9, 38.8] \\
Qwen3.5-35B & 13.3 [9.7, 17.3] & 5.0 [2.7, 7.7] & 6.3 [3.7, 9.3] & 61.7 [56.3, 66.7] & 36.6 [16.9, 56.6] \\
\bottomrule
\end{tabular}

%% file: tables/supplement_a_confidence_intervals.tex
\begin{tabular}{@{}lccc@{}}
\toprule
System & Typed@1 & Typed@3 & MRR \\
\midrule
GPT-5.2
  & 12.5 [6.9, 19.0] & 30.0 [21.4, 38.8] & .197 [.137, .256] \\
Claude 4.5
  & 14.2 [7.8, 20.5] & 32.5 [24.2, 41.2] & .221 [.160, .285] \\
Gemini 3 Flash
  & 23.3 [15.8, 31.4] & 41.7 [32.3, 51.6] & .310 [.232, .390] \\
GLM-4.6V
  & 10.0 [5.1, 15.9] & 27.5 [19.5, 36.1] & .174 [.116, .233] \\
Kimi K2.5
  & 15.0 [8.8, 21.8] & 32.5 [24.0, 41.2] & .228 [.164, .297] \\
Mistral 3.5
  & 15.8 [9.2, 23.0] & 25.8 [17.9, 34.2] & .200 [.133, .276] \\
Llama 4 Maverick
  & 14.2 [8.2, 20.7] & 26.7 [18.9, 34.7] & .192 [.130, .260] \\
Qwen3.5-35B
  & 14.2 [8.1, 20.9] & 24.2 [16.2, 31.9] & .190 [.127, .263] \\
\bottomrule
\end{tabular}

%% file: tables/supplement_p_significance_counts.tex
\begin{tabular}{@{}lrr@{}}
\toprule
Metric & Significant pairs & All pairs \\
\midrule
L0 & 17 & 28 \\
L1 & 18 & 28 \\
L2-Z & 16 & 28 \\
L2-E & 12 & 28 \\
L3 pair pass & 0 & 28 \\
\bottomrule
\end{tabular}